\documentclass[
11pt, 
english, 
singlespacing, 
headsepline, 
]{article}
\usepackage[super,comma,sort&compress]{natbib} 
\usepackage[nottoc]{tocbibind} 
\usepackage{graphicx}
\usepackage[margin=1.5in]{geometry}
\usepackage[shortlabels]{enumitem}
\usepackage{tipa}
\usepackage{amsmath}
\usepackage{xcolor}
\usepackage{array}
\usepackage{textgreek}
\usepackage[utf8]{inputenc}
\usepackage[T1]{fontenc}
\usepackage{listings}
\usepackage[titletoc]{appendix}

\title{\huge An efficient automated data analytics approach to large scale computational comparative linguistics \\
\vspace*{1cm}
}
\author{\begin{tabular}{c}Gabija Mikulyte\\{\small\it gabija.mikulyte@gmail.com}\end{tabular}  and 
\begin{tabular}{c}David Gilbert\\{\small\it david.gilbert@brunel.ac.uk}\end{tabular}\\ \\
Department of Computer Science\\ Brunel University London\\ Uxbridge UB8 3PH\\ U.K.}

\date{}
	
\begin{document}

\pagenumbering{gobble}
\maketitle
\newpage
\tableofcontents
\newpage
\listoffigures
\newpage
\listoftables
\newpage
\pagenumbering{arabic}

\begin{abstract}

This research project aimed to overcome the challenge of analysing
human language relationships, facilitate the grouping of languages
and formation of genealogical relationship between them by developing
automated comparison techniques. Techniques were based on the
phonetic representation of certain key words and concept. Example
word sets included numbers 1-10 (curated), large database of numbers
1-10 and sheep counting numbers 1-10 (other sources), colours
(curated), basic words (curated).

To enable comparison within the sets the measure of Edit distance
was calculated based on Levenshtein distance metric. This metric
between two strings is the minimum number of single-character edits,
operations including: insertions, deletions or substitutions. To
explore which words exhibit more or less variation, which words are
more preserved and examine how languages could be grouped based on
linguistic distances within sets, several data analytics techniques
were involved. Those included density evaluation, hierarchical
clustering, silhouette, mean, standard deviation and Bhattacharya
coefficient calculations. These techniques lead to the development
of a workflow which was later implemented by combining Unix shell
scripts, a developed R package and SWI Prolog. This proved to be
computationally efficient and permitted the fast exploration of large language sets
and their analysis.

\end{abstract}

\section{Introduction}
The need to uncover presumed underlying linguistic evolutionary principles and analyse correlation between world's languages has entailed this research. For centuries people have been speculating about the origins of language, however this subject is still obscure. Non-automated linguistic analysis of language relationships has been complicated and very time-consuming. Consequently, this research aims to apply a computational approach to compare human languages. It is based on the phonetic representation of certain key words and concept. This comparison of word similarity aims to facilitate the grouping of languages and the analysis of the formation of genealogical relationship between languages. 

This report contains a thorough description of the proposed methods, developed techniques and discussion of the results. During this projects several collections of words were gathered and examined, including colour words and numbers. The methods included edit distance, phonetic substitution table, hierarchical clustering with a cut and other analysis methods. They all aimed to provide an insight regarding both technical data summary and its visual representation.

\section{Background}
    \subsection{Human languages}
    For centuries, people have speculated over the origins of language and its early development. It is believed that language first appeared among Homo Sapiens somewhere between 50,000 and 150,000 years ago  \cite{humanLang2}. However, the origins of human language are very obscure.
    
    To begin with, it is still unknown if the human language originated from one original and universal Proto-Language. Alfredo Trombetti made the first scientific attempt to establish the reality of monogenesis in languages. His investigation concluded that it was spoken between 100,000 and 200,000 years ago, or close to the first emergence of Homo Sapiens \cite{trombetti}. However it was never accepted comprehensively. The concept of Proto-Language is purely hypothetical and not amenable to analysis in historical linguistics. 
    
    Furthermore, there are multiple theories of how language evolved. These could be separated into two distinctly different groups. 
    
    Firstly, some researchers claim that language evolved as a result of other evolutionary processes, essentially making it a by-product of evolution, selection for other abilities or as a consequence of yet unknown laws of growth and form. This theory is clearly established in Noam Chomsky \cite{chomsky} and Stephen Jay Gould's work \cite{gould}. Both scientists hypothesize that language evolved together with the human brain, or with the evolution of cognitive structures. They were used for tool making, information processing, learning and were also beneficial for complex communication. This conforms with the theory that as our brains became larger, our cognitive functions increased.
    
    Secondly, another widely held theory is that language came about as an evolutionary adaptation, which is when a population undergoes a change in process over time to survive better. Scientists Steven Pinker and Paul Bloom in ``Natural Language and Natural Selection'' \cite{humanLang1} theorize that a series of calls or gestures evolved over time into combinations, resulting in complex communication.
    
    Today there are 7,111 distinct languages spoken worldwide according to the 2019 Ethnologue language database. Many circumstances such as the spread of old civilizations, geographical features, and history determine the number of languages spoken in a particular region. Nearly two thirds of languages are from Asia and Africa. 
    
    The Asian continent has the largest number of spoken languages - 2,303. Africa follows closely with 2,140 languages spoken across continent. However, given the population of certain areas and colonial expansion in recent centuries, 86 percent of people use languages from Europe and Asia. It is estimated that there is around 4.2 billion speakers of Asian languages and around 1.75 billion speakers of European languages. 
    
    Moreover, Pacific languages have approximately 1,000 speakers each on average, but altogether, they represent more than a third of our world’s languages. Papua New Guinea is the most linguistically diverse country in the world. This is possibly due to the effect of its geography imposing isolation on communities. It has over 840 languages spoken, with twelve of them lacking many speakers. It is followed by Indonesia, which has 709 languages spoken across the country.

        \subsubsection{Indo-European languages and Kurgan Hypothesis}
        Indo-European languages is a language family that represents most of the modern languages of Europe, as well as specific languages of Asia. Indo-European language family consist of several hundreds of related languages and dialects. Consequently, it was an interest of the linguists to explore the origins of the Indo-European language family. 
        
\begin{figure}[htb]
            \centering
            \includegraphics[width=0.8\textwidth]{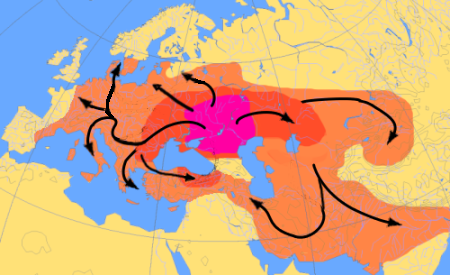}
            \caption{Indo-European migrations \cite{kurganH}}
            \label{kurgan}
        \end{figure}

        In the mid-1950s, Marija Gimbutas, a Lithuanian-American archaeologist and anthropologist, combined her substantial background in linguistic paleontology with archaeological evidence to formulate the Kurgan hypothesis \cite{gimbutas}. This hypothesis is the most widely accepted proposal to identify the homeland of Proto-Indo-European~(PIE) (ancient common ancestor of the Indo-European languages) speakers and to explain the rapid and extensive spread of Indo-European languages throughout Europe and Asia \cite{mallory} \cite{marler}. The Kurgan hypothesis proposes that the most likely speakers of the Proto-Indo-European language were people of a Kurgan culture in the Pontic steppe, by the north side of the Black Sea. It also divides the Kurgan culture into four successive stages (I, II, III, IV) and identifies three waves of expansions (I, II, III). In addition, the model suggest that the Indo-European migration was happening from 4000 to 1000 BC. See figure \ref{kurgan} for visual illustration of Indo-European migration.
        
\begin{figure}[htb]
            \centering
            \includegraphics[width=1\textwidth]{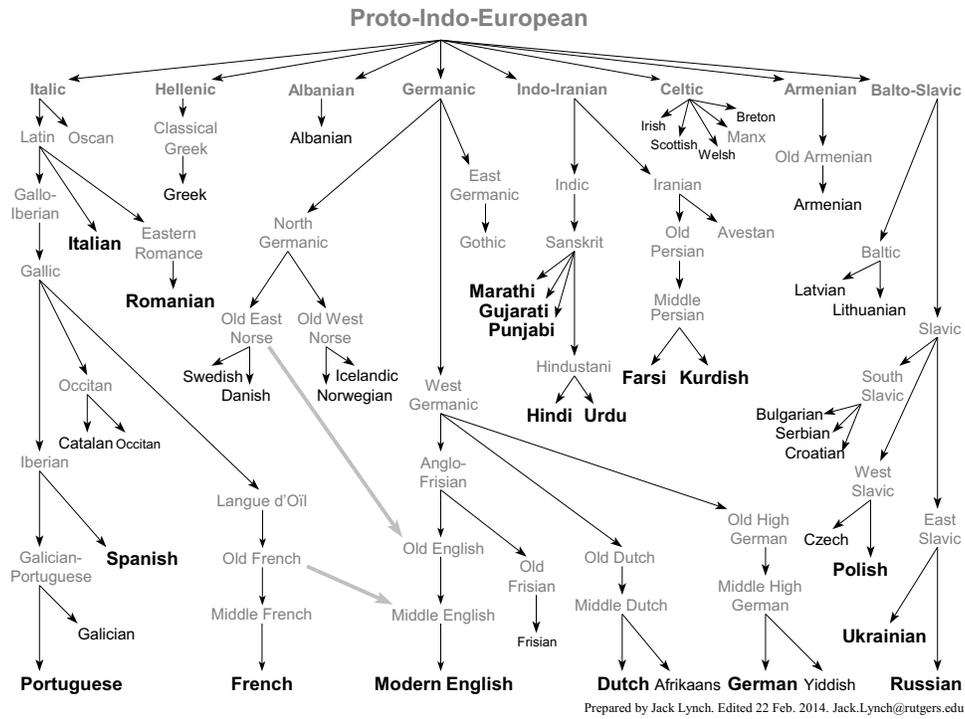}
            \caption[Indo-European language tree]{Indo-European language tree \cite{langTree}}
            \label{IElanguagetree}
        \end{figure}

        Today there are approximately 445 living Indo-European languages, which are spoken by 3.2 billion people, according to Ethnologue. They are divided into the following groups: Albanian, Armenian, Baltic, Slavic, Celtic, Germanic, Hellenic, Indo-Iranian and Italic~(Romance) \ref{IElanguagetree} \cite{langTree}.
        
        \subsubsection{Brittonic languages}
        Brittonic or British Celtic languages derive from the Common Brittonic language, spoken throughout Great Britain south of the Firth of Forth during the Iron Age and Roman period. They are classified as Indo-European Celtic languages \cite{brit}. The family tree of Brittonic languages is showed in Table \ref{brittonic}. Common Brittonic is ancestral to Western and Southwestern Brittonic. Consequently, Cumbric and Welsh, which is spoken in Wales, derived from Western Brittonic. Cornish and Breton, spoken in Cornwall and Brittany, respectively, originated from Southwestern side.  
        
        Today Welsh, Cornish and Breton are still in use. However, it is worth to point out that Cornish is a language revived by second-language learners due to the last native speakers dying in the late 18th century. Some people claimed that the Cornish language is an important part of their identity, culture and heritage, and a revival began in the early 20th century. Cornish is currently a recognised minority language under the European Charter for Regional or Minority Languages. 
        
        \begin{table}[h!]
        \centering
        \caption{Table of Brittonic languages}
        \label{brittonic}
             \begin{tabular}{|c|c|c|c|} 
                 \hline
                 \multicolumn{4}{|c|}{Common Brittonic} \\  
                 \hline\hline
                 \multicolumn{2}{|c|}{Western Brittonic} & \multicolumn{2}{|c|}{Sothwestern estern Brittonic} \\ 
                 \hline\hline
                 Cumbric & Welsh & Cornish & Breton \\
                 \hline
            \end{tabular}
        \end{table}
        
        \subsubsection{Sheep Counting System}
        Brittonic Celtic language is an ancestor to the number names used for sheep counting \cite{distinCultural} \cite{yanTanTethera}. Until the Industrial Revolution, the use of traditional number systems was common among shepherds, especially in the fells of the Lake District. The sheep-counting system was referred to as Yan Tan Tethera. It was spread across Northern England and in other parts of Britain in earlier times. The number names varied according to dialect, geography, and other factors. They also preserved interesting indications of how languages evolved over time.

        The word ``yan'' or ``yen'' meaning ``one'', in some northern English dialects represents a regular development in Northern English \cite{leithSocial}. During the development the Old English long vowel /\textipa{A:}/ <\={a}> was broken into /ie/, /ia/ and so on. This explains the shift to ``yan'' and ``ane'' from the Old English \={a}n, which is itself derived from the Proto-Germanic ``*ainaz'' \cite{griffithsDictionary}.
        
        In addition, the counting system demonstrates a clear connection with counting on the fingers. Particularly after numbers reach 10, as the best known examples are formed according to this structure: 1 and 10, 2 and 10, up to 15, and then 1 and 15, 2 and 15, up to 20. The count variability would end at 20. It might be due to the fact, that the shepherds, on reaching 20, would transfer a pebble or marble from one pocket to another, so as to keep a tally of the number of scores.



\section{Aims and Objectives}
    \subsection{Overall Aim} 
     The aim of this research was to develop computational methods to compare human languages based on the phonetic form of single words (i.e. not exploiting grammar). This comparison of word similarity aims to facilitate the grouping of languages, the identification of the the presumed underlying linguistic evolutionary principles and the analysis of the formation of genealogical relationship between languages.

    \subsection{Specific Objectives}
    \begin{enumerate}
      \item Devise a way to encode the phonetic representation of words, using: 
      \begin{enumerate}
        \item an in-house encoding,
        \item an IPA (International Phonetic Alphabet).
      \end{enumerate}
      \item Develop methods to analyze the comparative relationships between languages using: descriptive and inferential statistics, clustering, visualisation of the data, and analysis of the results.
      \item Implement a repeatable process for running the analysis methods with new data.
      \item Analyse the correlation between geographical distance and language similarity (linguistic distance), and investigate if it explains the evolutionary distance.
      \item Examine which words exhibit more or less variation and the likely causes of it.
      \item Explore which words are preserved better across the same language group and possible reasons behind it.
      \item Explore which language group preserves particular words more in comparison to others and potential reasons behind it.
      \item Determine if certain language groups are correct and exploit the possibility of forming new ones.
    \end{enumerate}

\section{Data}
    \subsection{Language files}\label{langFiles}
    Language file or database is a set of languages, each of which is associated with an ordered list of words. All lists of words for a particular data set have the same length. For example:
    
    \begin{small}
    \begin{verbatim}
numbers(romani,[iek,dui,trin,shtar,panj,shov,efta,oksto,ena,desh]).
numbers(english,[wun,too,three,foor,five,siks,seven,eit,nine,ten]).
numbers(french,[un,de,troi,katre,sink,sis,set,wuit,neuf,dis]).
    \end{verbatim}
    \end{small}

    Words and languages are encoded in this format for later use of Prolog. In Prolog each ``numbers'' line is a fact, which has 2 arguments; the first is the language name and the second is a list (indicated in between square brackets) of words. Words can be written down in their original form or encoded phonetically (as shown in the example). Where synonyms for a word are known, then the word itself is represented by a list of the synonym words.  In the example below, Lithuanian, Russian and Italian have two words for the English `blue':
    
    \begin{small}
    \begin{verbatim}
words(english,[black,white,red,yellow,blue,green]).
words(lithuanian,[juoda,balta,raudona,geltona,[melyna,zhydra],zhalia]).
words(russian,[chornyj,belyj,krasnyj,zholtyj,[sinij,goluboj],zeljonyj]).
words(italian,[nero,bianco,rosso,giallo,[blu,azzurro],verde]).
    \end{verbatim}
    \end{small}

    The main focus of this research was exploring words phonetically. Consequently, special encoding was used. It consisted of encoding phonemes by using only one letter; incorporating capital letters for encoding different sounds {(See table \ref{phonetic})}.
    
    \begin{table}[h!]
    \centering
    \caption{Table of phonetic encoding }
    \label{phonetic}
         \begin{tabular}{|c|c|} 
             \hline
             Symbol & Meaning \\ [0.5ex] 
             \hline\hline
             c & ts \\ 
             \hline
             x & ks \\
             \hline
             C & ch as in charity \\
             \hline
             k & as in cat \\
             \hline
             T & th \\ 
             \hline
             S & sh \\
             \hline
             G & dzh \\ 
             \hline
             K & kh \\
             \hline
             Z & zh \\ 
             \hline
             D & dz \\
             \hline
             H & Spanish/Portuguese sound of ``j'' \\
             \hline
             A, I, O, U & long vowels \\
             \hline
        \end{tabular}
    \end{table}

    Table \ref{databases} summarises the language files that are obtained at the moment.
        \begin{table}[h!]
        \caption{Table of Language files}
        \label{databases}
        \centering
         \begin{tabular}{|m{4cm}|m{2cm}|m{2cm}|m{4cm}|} 
             \hline
             Name & Number of languages & Number of words per language & Description \\ [0.5ex] 
             \hline\hline
             Numbers Small Collection & 92 & 10 & Numbers 1 to 10 \\ 
             \hline
             Numbers Big Collection & 3880 & 10 & Numbers 1 to 10 \\
             \hline
             Sheep Counting Numbers & 54 & 10 & Numbers 1 to 10 \\
             \hline
             Basic Words & 42 & 13 & Concept words: sun, moon, rain, water, fire, man, woman, mother, father, child, yes, no, blood \\ 
             \hline
             Colours & 23 & 6 & Main colours: black, white, red, yellow, blue, green \\
             \hline
             Basic Words IPA & 3 & 12 & Concept words as in ``Basic Words'' \\
             \hline
             Numbers IPA & 3 & 10 & Numbers 1 to 10\\ 
             \hline
             Colours IPA & 3 & 6 & Colours as in ``Colours'' \\
             \hline
        \end{tabular}
    \end{table}
    
    \subsection{Sheep}
    \subsubsection{Sheep counting words}
    \par Sheep counting numbers were extracted from ``Yan Tan Tethera'' \cite{yanTanTethera} page on Wikipedia and placed in a Prolog database. Furthermore, data was encoded phonetically using the set of rules provided by Prof.~David Gilbert.
    \par In the given source, number sets ranged from 1-3 to 1-20 for different dialects. The initial step was to reduce the size of the data to sets of numbers 1-10. This way aiming:
    \begin{enumerate}[(a)]
        \item to have Prolog syntax without errors (avoided ``-'', ``~'' as they were common symbols after numbers reached 10);
        \item to avoid the effects of different methods of forming and writing down numbers higher than 10. (Usually they were formed from numbers 1-10 and a base. However, they were written in a different order, making the comparison inefficient.)
    \end{enumerate}
    \par In addition, the Wharfedale dialect was removed since only numbers 1-3 were provided; the Weardale dialect was eliminated as it had a counting system with base 5. 
     Consequently, the final version of sheep counting numbers database consisted of 23 observations (dialects) with numbers 1-10.
     \subsubsection{Geographical data}
\begin{figure}[htb]
         \centering
         \includegraphics[width=0.8\textwidth]{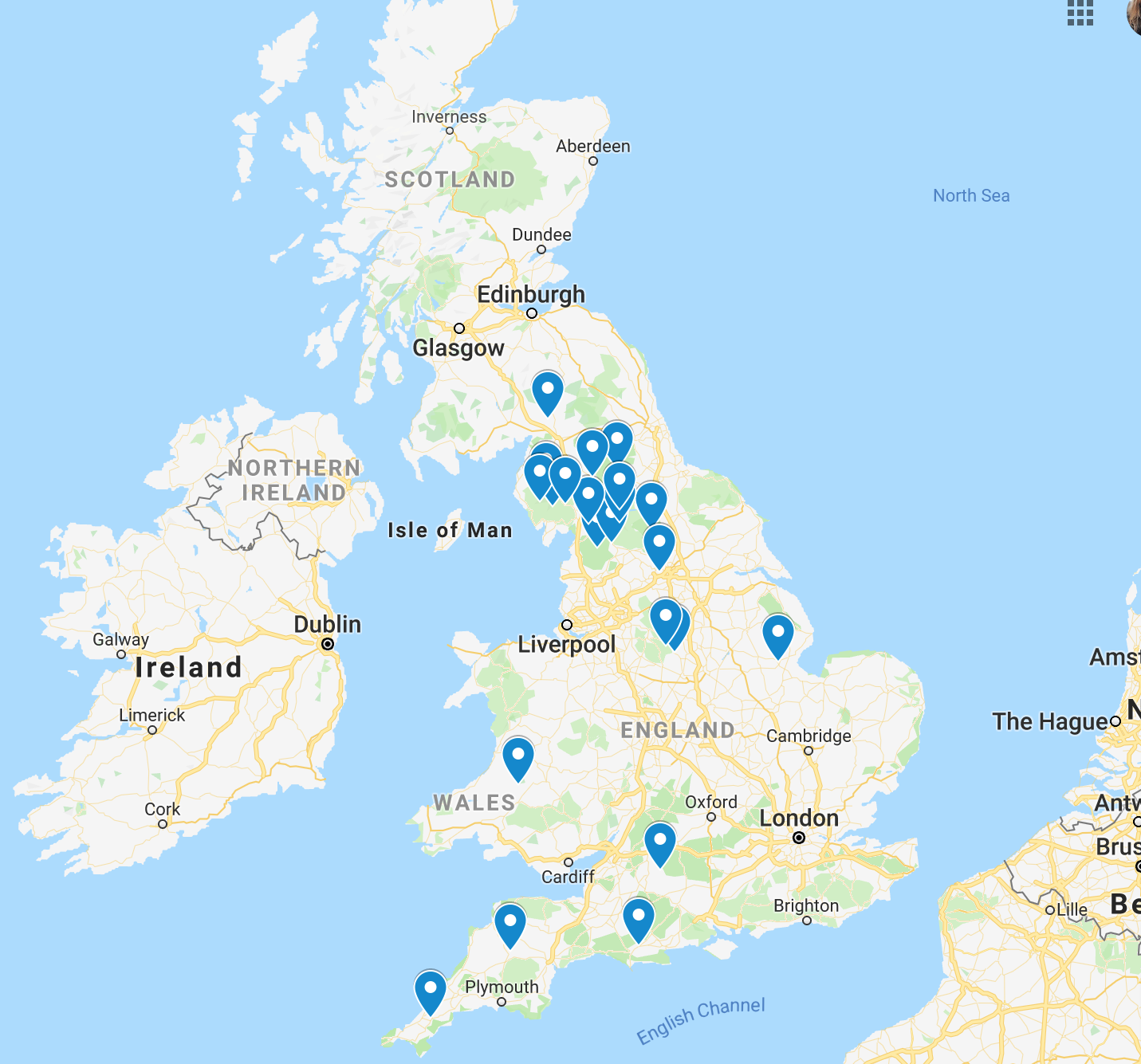}
         \caption[Sheep dialects in Britain]{Sheep dialects in Britain.  A map with 23 pins, noting the places of different dialects}
         \label{gMap}
     \end{figure}

     In order to enable the analysis of linguistic and geographical distance relationship, a geographical distance database was created. It was done by firstly creating a personalized Google Map with 23 pins, noting the places of different dialects {(they were located approximately in the middle of the area)} {(Figure: \ref{gMap})}. Subsequently, pairwise distances were calculated between all of them {(taking walking distance)} and added to the database for further use.
     
    \subsection{Colours}
    Colour words were extracted from ``Colour words in many languages'' \cite{omniglot} page on Omniglot, collected from people and dictionaries. In addition, data was encoded phonetically using the set of rules provided by Prof.~David Gilbert. 
    
    The latest version of the database consisted of 42 different languages, each containing 6 colours: black, white, red, yellow, blue, green. For the purposes of analysis the following groups were created:
    \begin{enumerate}[(a)]
        \item All languages - ``ColoursAll'' (42 languages)
        \item Indo-European languages - ``ColoursIE'' (39 languages)
        \item Germanic languages - ``ColoursPGermanic'' (10 languages)
        \item Romance languages - ``ColoursPRomance'' (11 languages)
        \item Germanic and Romance languages - ``ColoursPG\_R'' (21 languages)
    \end{enumerate}

    \subsection{IPA}
    ``Automatic Phonemic Transcriber'' \cite{IPA} was used to create 3 IPA encoded databases:
    \begin{enumerate}[(a)]
        \item ``BasicWords'' - words in their original form were taken from Prof.~David Gilbert's database for basic words (including: sun, moon, rain, water, fire, man, woman, mother, father, child, yes, no, blood).
        \item ``Numbers'' - numbers from 1-10 in their original form were taken from Prof.~David Gilbert's small database of numbers.
        \item ``Colours'' - words were taken from the above mentioned database (including words: black, white, red, yellow, blue, green).
    \end{enumerate}
    Each of the above mentioned databases consisted of 3 languages: English, Danish and German (these were the languages the Automatic Phonemic Transcriber provided) all encoded in IPA.
    
    As the research progressed, the difficulty of obtaining IPA encoding for different languages was faced. This study could not find a cross-linguistic IPA dictionary that included more than 3 languages. Consequently, the question of its existence was raised.
    
\section{Methodology}
There are two main processes to be carried out. 

\begin{figure}[htb]
    \centering
    \includegraphics[width=0.6\textwidth]{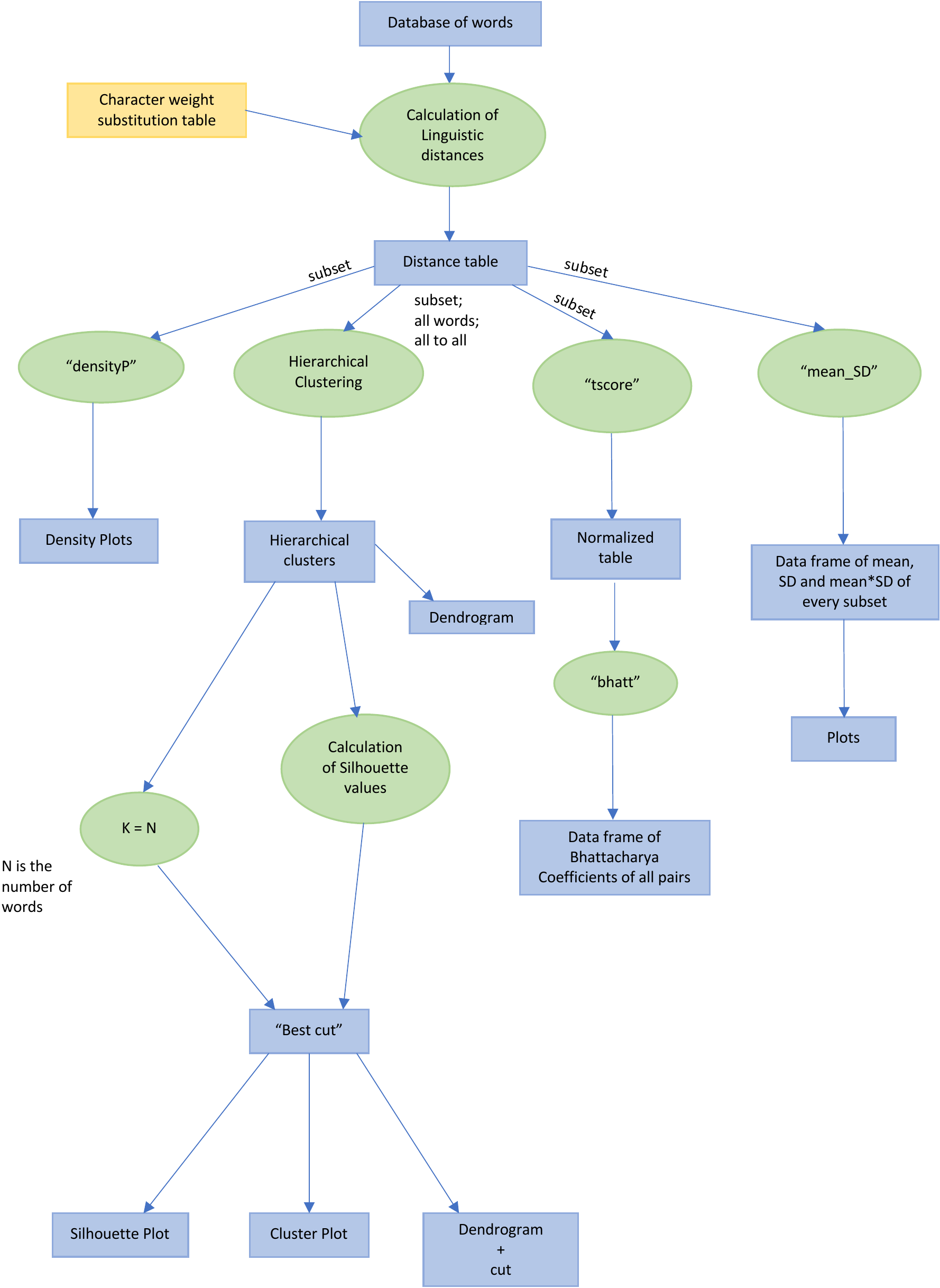}
    \caption{Work-flow of comparison of languages}
    \label{workflow_words}
\end{figure}

The first process {(Figure: \ref{workflow_words})} aims to analyse a databases of words; explore which words exhibit more or less variation, which words are more preserved; examine how languages could be grouped based on linguistic distances of words. 

It begins with the calculation of pairwise linguistic distances for the given database of words. A Phonetic Substitution Table is used to assign weights during the calculation and could possibly be modified. The result is a new distance table which is analysed in the following ways:
\begin{itemize}
    \item Performing ``densityP'' function. The outcome is density plots for every word of a database.
    \item Performing Hierarchical clustering. After, the ``Best cut'' is determined, which is either the best Silhouette value after calculation of all possible cases, or a forced number K which is a number of words per language in the language file
    \item Calculating Bhattacharya coefficients.
    \item Performing ``mean\_SD'' function.
\end{itemize}

\begin{figure}[htb]
    \centering
    \includegraphics[width=0.8\textwidth]{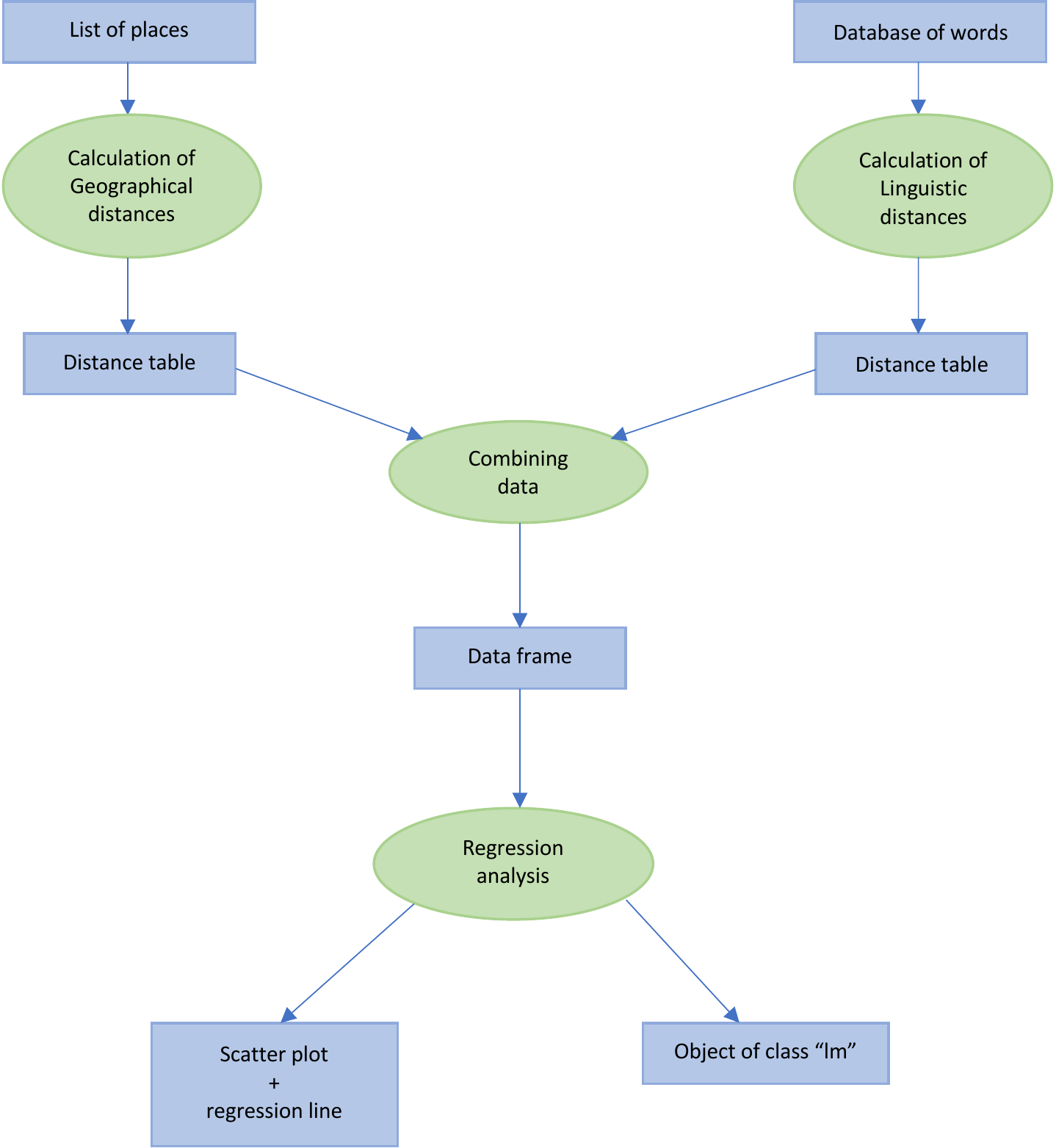}
    \caption{Work-flow of relationship analysis}
    \label{workflow_relationship}
\end{figure}

The second process {(Figure: \ref{workflow_relationship})} targets to investigate the relationship between two sets of distance data. In this research, it was applied to analyse the relationship between linguistic and geographical distances.

It starts with producing two pairwise distance tables: one of them is calculated geographical distances, another one is calculated linguistic distances. Then the data from both tables is combined into a data frame for regression analysis in R. The outcome is an object of the class ``lm'' (result of R function ``lm'' being used), that is used for data analysis, and a scatter plot with a regression line for visual analysis.

Both processes have been automated, see Section~\ref{automation}.

\section{Methods}\label{methods}
   
    \subsection{Edit Distance}
     For the purposes of this research Edit distance (a measure in computer science and computational linguistics for determining the similarity between 2 strings) was calculated based on Levenshtein distance metric. This metric between two strings is the minimum number of single-character edits, operations including: insertions, deletions or substitutions. 
    
    The Levenshtein distance between two strings a,b (of length $\mid a\mid$ and $\mid b\mid$ respectively) is given by $lev_{a,b}(\mid a \mid , \mid b \mid)$ where 
      $$
      lev_{a,b}(i,j)= 
      \begin{cases}
       max(i,j) & \text{\hspace{15.5em}If min$(i,j)$=0}\\
       min & \begin{cases}
                lev_{a,b}=(i-1,j) + 1\\
                lev_{a,b}=(i,j-1) + 1 ,\hspace{1.9em}& \text{otherwise}\\
                lev_{a,b}= (i-1,j-1) + 1_{(a_{i}\neq b_{j})}
             \end{cases}
       \end{cases}
       $$
    \\ where $1_{(a_{i}\neq b_{j})}$ is the indicator function equal to 0 when $a_{i}=b_{j}$ and equal to 1 otherwise.  A {\it normalised} edit distance between two strings can be computed by 
    $$lev\_norm_{a,b} = \dfrac{lev_{a,b}}{max(\mid a \mid , \mid b \mid)}$$
    
    Edit distance was implemented by Prof.~David Gilbert using dynamic programming in SWI Prolog~\cite{wielemaker:2011:tplp}. The program was used to compare two words with the same meaning from different languages. When pairwise comparing two words where either one or both comprise synonyms, all the alternatives for each word one one language are compared with the corresponding (set) of words in the other language, and the closest match is selected.  In addition, all to all comparisons were made, i.e. edit distance was calculated for words having different meaning as well.  Finally, the edit distance for two languages represented by two lists of equal length of corresponding words was computed by taking the average of the edit distance for each (corresponding) pair of words.

    An example of pairwise alignments is for the pair of words
overa-hofa, where 3 alignments are produced with the use of gap penalty $=1$ and substitution penalties $f \leftrightarrow v = 0.2$, $e \leftrightarrow o = 0.2$ and all other mismatches 1:
    \begin{verbatim}
[[-,h],[o,o],[v,f],[e,-],[r,-],[a,a]]
[[o,-],[v,h],[e,o],[r,f],[a,a]]
[[o,h],[v,-],[e,o],[r,f],[a,a]]
    \end{verbatim}
    each with the raw edit distance of 3.2, and the normalised edit distance of 
    $$\dfrac{3.2}{max(\mid \mbox{\tt overa} \mid, \mid \mbox{\tt hofa} \mid )} = \dfrac{3.2}{5} = 0.64$$
    
    For the sake of clarity we can write the first alignment for example as
    \begin{center}
    \begin{tabular}{cccccc}
    - & o & v & e & r & a \\
    h & o & f & - & - & a
    \end{tabular}
    \end{center}
    where only 3 letters are directly aligned.
    
    \subsection{Phonetic Substitution Table}
     In order to give a specified weight for different operations (insertion, deletion and substitution) Phonetic Substitution Table was created by incorporating Grimm's law \cite{grimm} and extending it in-house.
     
\begin{figure}
        \centering
        \includegraphics[width=0.8\textwidth]{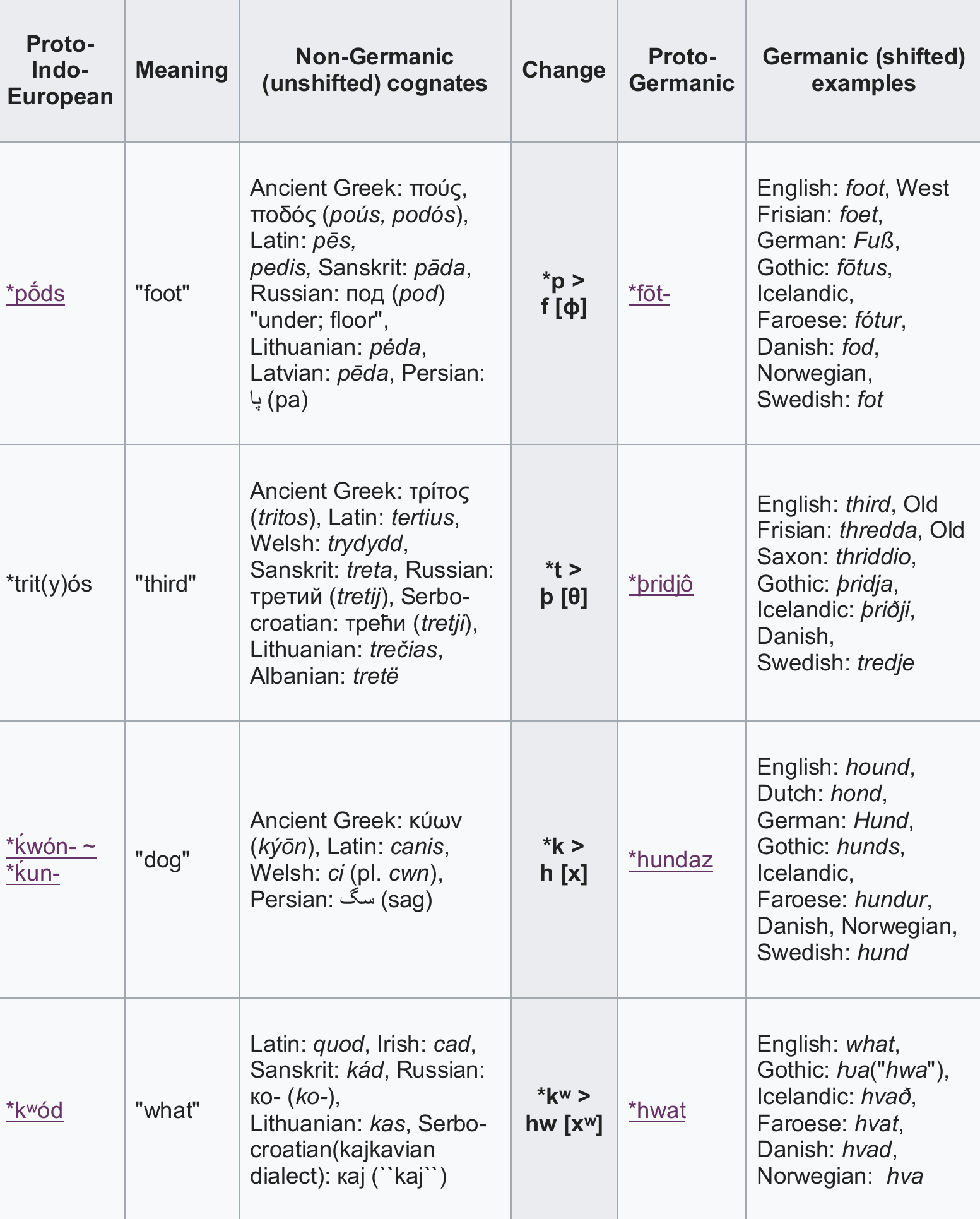}
        \caption{Table illustrating Grimm's Law chain shift}
        \label{grim}
    \end{figure}

     Grimm's Law, principle of relationships in Indo-European languages, describes a process of the regular shifting of consonants in groups. It consist of 3 phases in terms of chain shift \cite{grimmChain}.
    \begin{enumerate}
        \item Proto-Indo-European voiceless stops change into voiceless fricatives.
        \item Proto-Indo-European voiced stops become voiceless stops.
        \item Proto-Indo-European voiced aspirated stops become voiced stops or fricatives.
    \end{enumerate}
    This is an abstract representation of the chain shift:
    \begin{itemize}
    \centering
        \item[] $bh > b > p > $ \textipa{F}
        \item[] $dh > d > t > $ \textipa{T}
        \item[] $gh > g > k > x$
        \item[] $gwh > gw > kw > xw$
    \end{itemize}
    Figure \ref{grim} illustrates how further consonant shifting following Grimm's law affected words from different languages \cite{grimmy}.

    Phonetic substitution table was extended in-house by adding more shifts. In addition, it was also written in the way to work with the special encoding described in \ref{langFiles} section.
    Find the full table ``editable'' in Appendix~\ref{table}.

    Another phonetic substitution table, called ``editableGaby'', was made (See Appendix~\ref{table}). It was extended by adding pairs like ``dzh'' and ``zh''; ``dzh'' and ``ch''; ``kh'' and ``g''; as well as ``H''(sound of e.g. spannish/portuguese ``j'') with ``kh'', ``g'', ``k'', ``h''. In addition, some of the weights were changed for certain pairs for experimental purposes. 
    
    \subsection{Hierarchical Clustering}
    
        \subsubsection{Using the OC program}
        The OC program \cite{oc} is general purpose hierarchical cluster analysis program. It outputs a list of the clusters and optionally draws a dendrogram in PostScript. It requires complete upper diagonal distance or similarity matrix as an input.

        \subsubsection{Using R}\label{usingR}
        Hierarchical clustering in R was performed by incorporating clustering together with Silhouette value calculation and cut performance.
        
        In order to fulfill agglomerative hierarchical clustering more efficiently, we created a set of functions in R: 
        \begin{enumerate}
            \item ``sMatrix'' - Makes a symmetric matrix from a specified column. The function takes a specifically formatted data frame as an input and returns a new data frame. Having a symmetric matrix is necessary for ``silhouetteV'' and ``hcutVisual'' functions.
            \item ``silhouetteV'' - Calculates Silhouette values with ``k'' value varying from 2 to n-1 (n being the number of different languages/number of rows/number of columns in a data frame). The function takes a symmetric distance matrix as an input and returns a new data frame containing all Silhouette values.
            \item ``hcutVisual'' - Performs hierarchical clustering and makes a cut with the given K value. Makes Silhouette plot, Cluster plot and dendrogram. Returns a ``hcut'' object from which cluster assignment, silhouette information, etc. can be extracted.
        \end{enumerate}
    
    It is important to note that K-Means clustering was not performed as the algorithm is meant to operate over a data matrix, not a distance matrix.  
        
    \subsection{Further analysis with R}\label{R2}
    Another set of functions was created to analyse collected data further. They target to ease the comparison of the mean, standard deviation, Bhattacharya coefficient within the words or language groups. Including:
    \begin{enumerate}
        \item ``mean\_SD'' - Calculates mean, standard deviation, product of the mean and the SD multiplication for every column of the input. Visualises all three values for each column and places it in one plot, which is returned.
        \item ``densityP'' - Makes a density plot for every column of the input and puts it in one plot, which is returned.
        \item ``tscore'' - Calculates t-score for every value in the given data frame. (T-score is a standard score Z shifted and scaled to have a mean of 50 and a standard deviation of 10)
        \item ``bhatt'' - Calculates Bhattacharya coefficient (the probability of the two distributions being the same) for every pair of columns in the data frame. The function returns a new data frame.
    \end{enumerate}
    
    \subsection{Process automation}\label{automation}
    In order to optimise and perform analysis in the most time-efficient manner processes of comparing languages were automated. It was done by creating two shell scripts and an R script for each of them.
    
    The first shell script named ``oc2r\_hist.sh'' was made to perform hierarchical clustering with the best silhouette value cut. This script takes a language database as an input and performs pairwise distance calculation. It then calls ``hClustering.R'' R script, which reads in the produced OC file, performs hierarchical clustering and calculates all possible silhouette values. Finally, it makes a cut with the number of clusters, which provides the highest silhouette value. To enable this process the R script was written by incorporating the functions described in section \ref{usingR}. The outcome of this program is a table of clusters, a dendrogram, clusters' and silhouette plots.
    
    The second shell script called ``wordset\_make\_analyse.sh'' was made to perform calculations of mean, standard deviation, Bhattacharya scores and produce density plots. This script takes a language database as an input and performs pairwise distance calculations for each word of the database. It then calls ``rAnalysis.R'' R script, which reads in the produced OC file and performs further calculations. Firstly, it calculates mean, standard deviation and the product of both of each word and outputs a histogram and a table of scores. Secondly, it produces density plots of each word. Finally, it converts scores into T-Scores and calculates Bhattacharya coefficient for every possible pair of words. It then outputs a table of scores. To enable this process the R script was written by incorporating the functions described in section \ref{R2}.

    Finally, both of the scripts were combined to minimise user participation.

\section{Results}
    \subsection{Sheep}
    The sheep counting database was evaluated in the following ways:
    \begin{itemize}
    \item Obtaining average pairwise linguistic distance, pairwise linguistic distance of subsets (different words),
    \item Performing all to all comparison (where linguistic distance is calculated between words with different meaning, as well as with the same),
    \item Collecting geographical data and comparing relationship between linguistic and geographical distances.
    \end{itemize}
   Upon generation of the above mentioned data, the methods defined in \ref{methods} section were used.
    
        \subsubsection{Analysis of average and subset linguistic distance}
\begin{figure}[htb]
            \centering
            \includegraphics[width=0.8\textwidth]{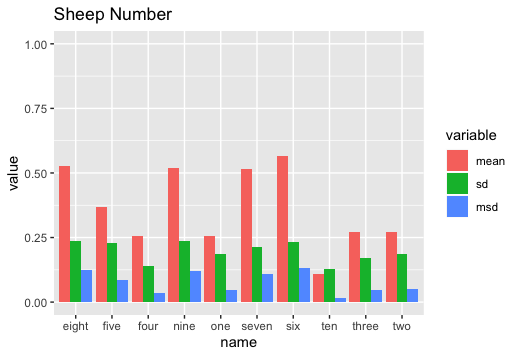}
            \caption[Statistics of sheep counting systems]{Mean, SD and mean*SD of every number of sheep counting systems}
            \label{SheepNumbers}
        \end{figure}

\begin{figure}[htb]
            \centering
            \includegraphics[width=0.8\textwidth]{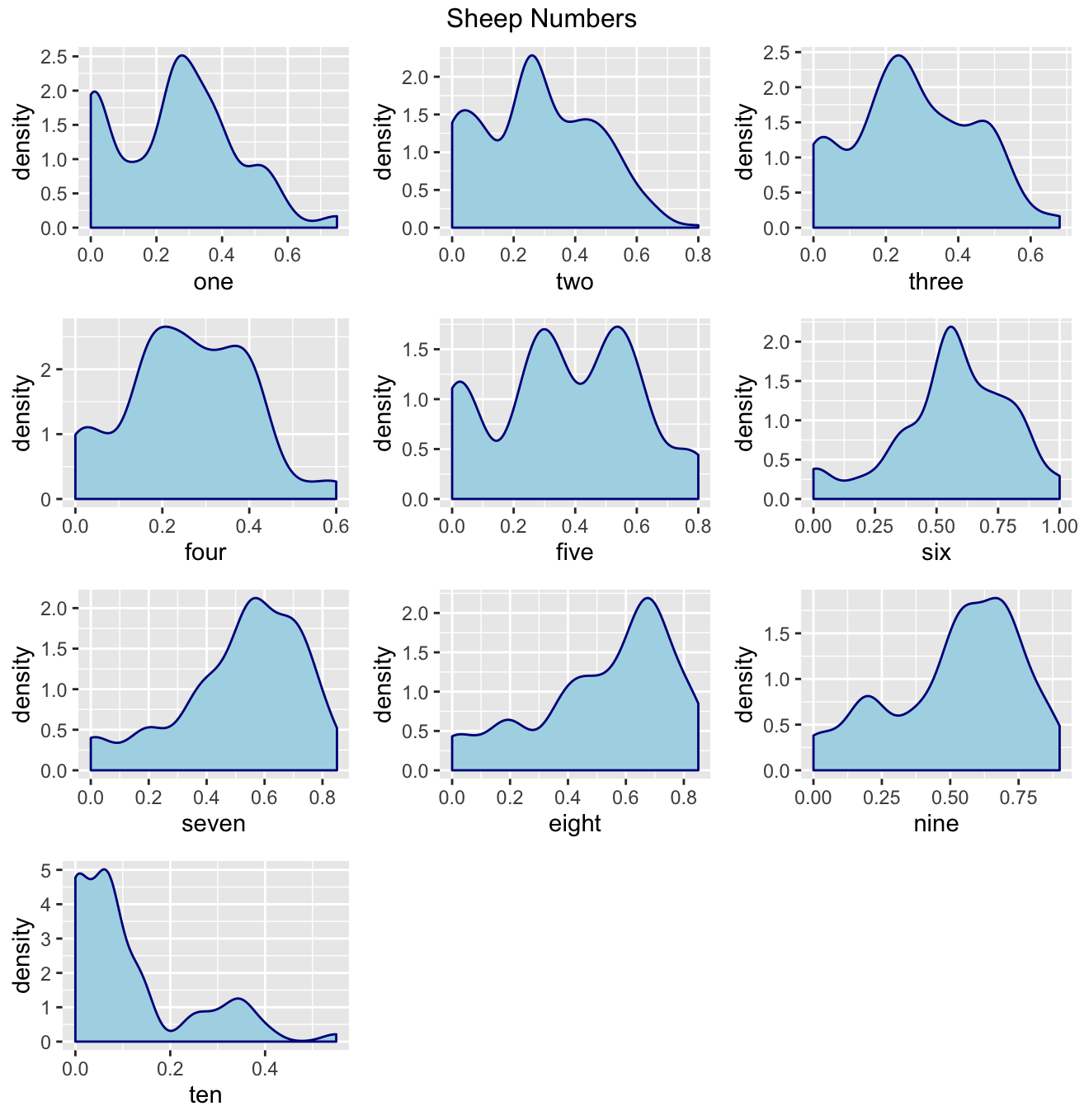}
            \caption[Density plot of sheep counting systems]{Density plots of each number of sheep counting systems}
            \label{SheepNumbersDistribution}
        \end{figure}

        After applying functions ``mean\_SD'' {(Figure: \ref{SheepNumbers})} and ``densityP'' {(Figure: \ref{SheepNumbersDistribution})} to the linguistic distances of every word (numbers 1 to 10) in R, the following observations were made. First of all, the most preserved number across all dialects was ``10'' with distance mean 0.109 and standard deviation 0.129. Numbers ``1'', ``2'', ``3'', ``4'' had comparatively small distances, which might be the result of being used more frequently. On the other hand, number ``6'' showed more dissimilarities between dialects than other numbers. The mean score was 0.567 and standard deviation - 0.234. The product scores of mean and standard deviation helped to evaluate both at the same time. Moreover, density plots showed significant fluctuation and tented to have a few peaks. But in general, conformed with the statistics provided by ``mean\_SD''.

        \begin{figure}[htb]
            \centering
            \includegraphics[width=0.8\textwidth]{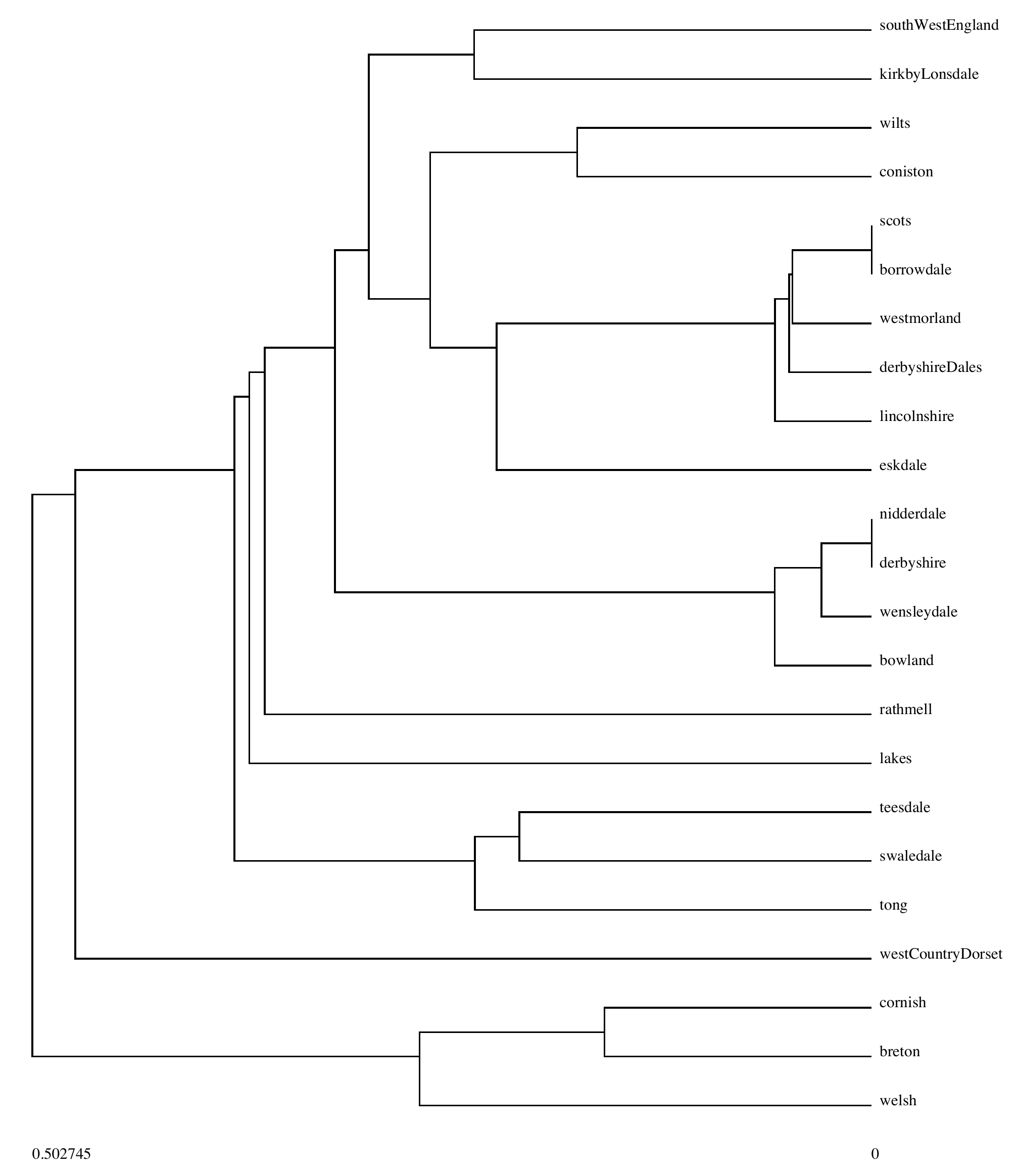}
            \caption[Dendrogram of sheep counting systems]{Dendrogram of linguistic distances of 23 dialects}
            \label{yttD}
        \end{figure}
        
        \subsection{Hierarchical clustering}
\begin{figure}
            \centering
            \includegraphics[width=0.8\textwidth]{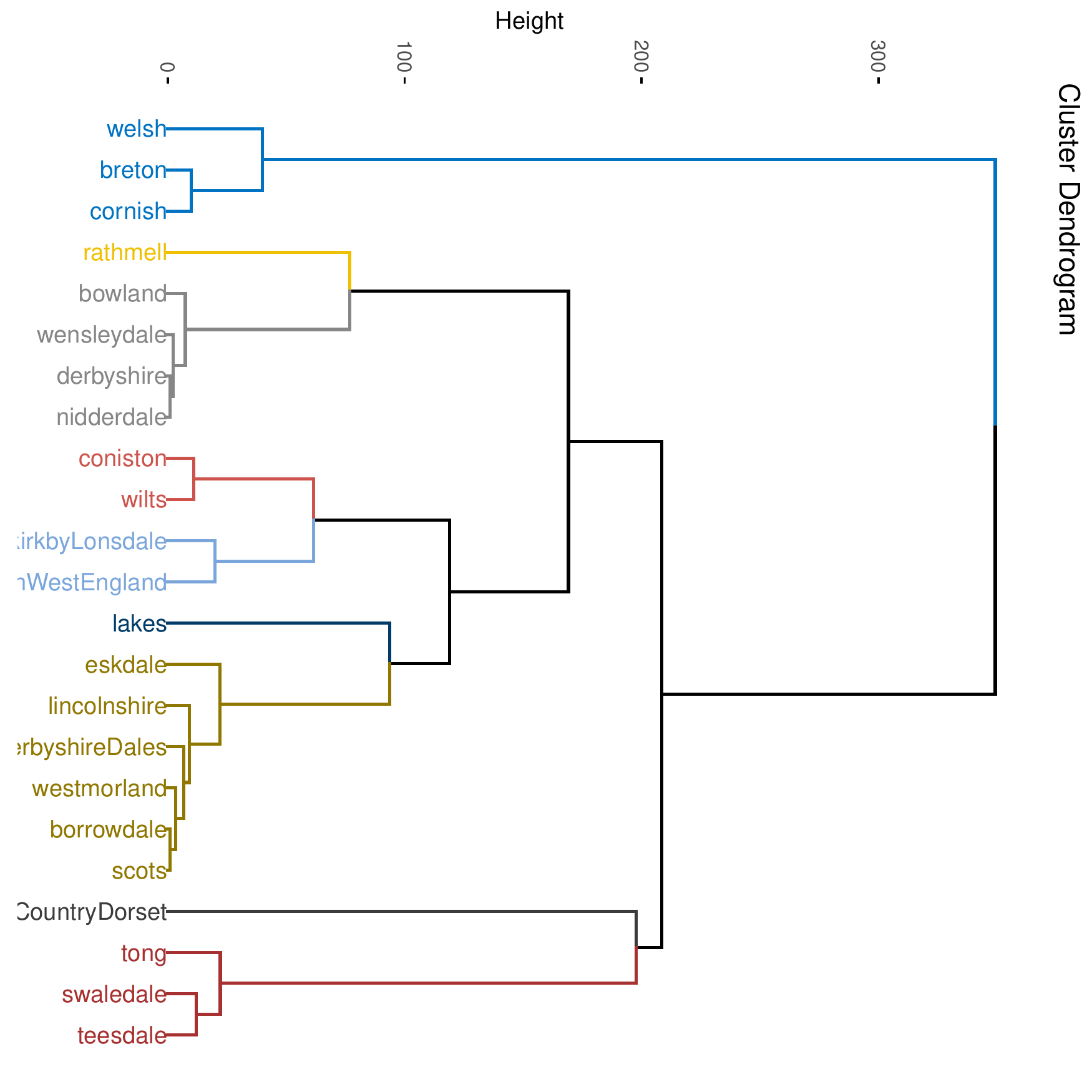}
            \caption{Dendrogram of sheep counting systems with the best Silhouette cut}
            \label{yttS}
        \end{figure}

        Hierarchical clustering was performed with the best Silhouette value cut (Figure \ref{yttS}). The Silhouette value suggested making 9 clusters. In this grouping, the most interesting observation was that Welsh, Breton and Cornish languages were placed together. It conforms with the fact that all 3 languages descended directly from the Common Brittonic language spoken throughout Britain before the English language became dominant.
        
        \subsubsection{All to all comparison analysis}
        To enable analysis of clusters of all to all comparison, hierarchical clustering was performed. This was done by two different approaches: calculating a silhouette value and choosing the number of clusters accordingly; forcing a function to make 10 clusters due to having numbers from 1 to 10 in the sheep counting database.
    
        By using function ``silhouetteV'' silhouette values were calculated for all possible \(k\) values. The returned data frame indicated the best number of clusters being 70 {(see Appendix~\ref{Asheep} for dendrogram and cluster plot)}. The suggested clusters were not distinguished with very high clarity in terms of numbers 1-10 perfectly, but they were comparatively good. A pattern that numbers, which had lower mean and standard deviation scores, would result in purer clusters was noticed. Clusters of numbers ``1'', ``2'', ``3'', ``4'', ``5'' and ``10'' were not as mixed as ``6'', ``7'', ``8'', ``9''.
        
\begin{figure}[htb]
            \centering
            \includegraphics[width=0.8\textwidth]{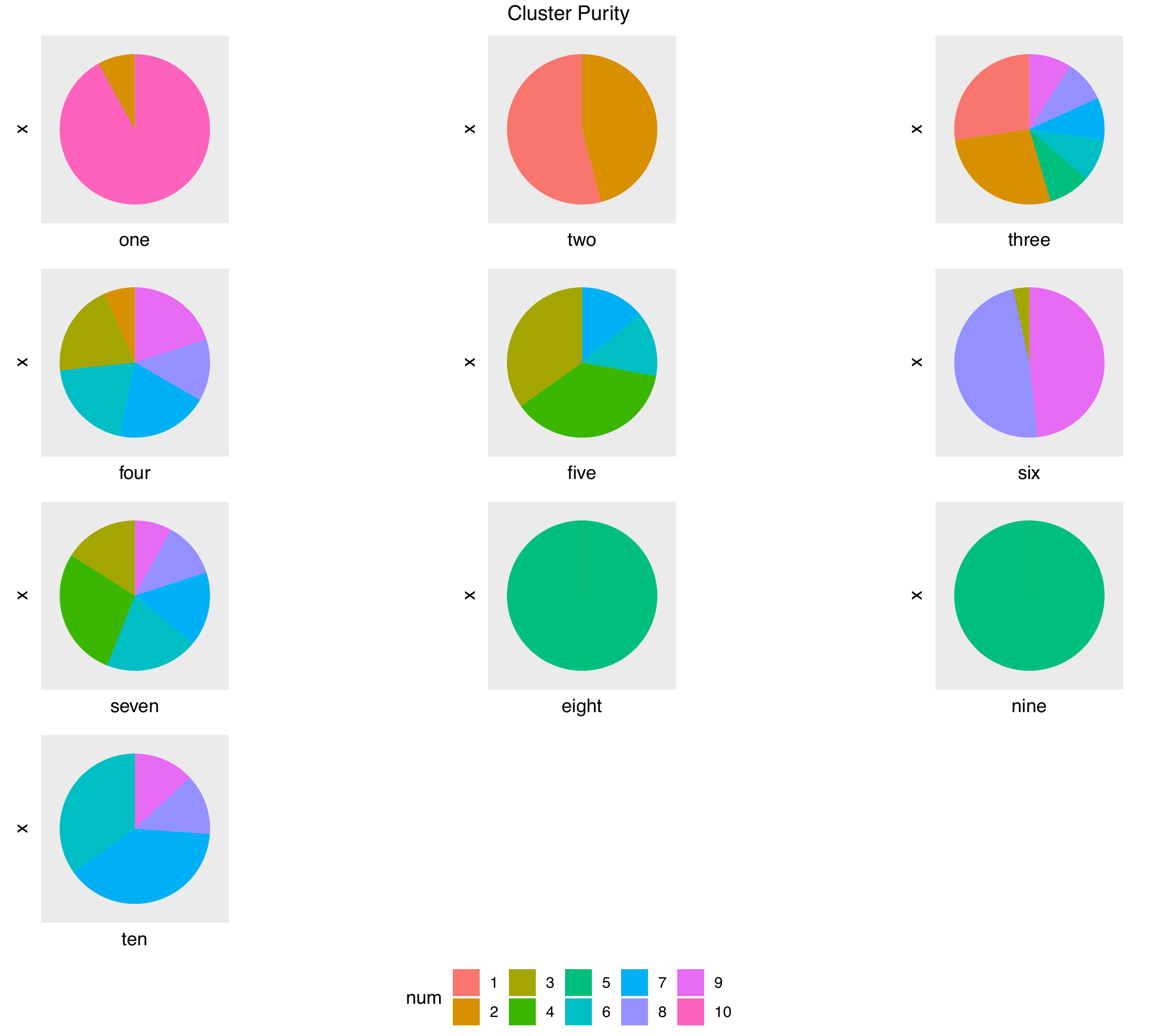}
            \caption[Purity of clusters of sheep counting systems]{Purity of hierarchical clusters of sheep counting systems all to all comparison.  Clusters numbered according to $K=10$.  Colour key indicates number words. }
            \label{sheepaxaCP}
        \end{figure}

        Another way of looking at all to all comparison data was by producing 10 clusters. It was done by using ``hcutVisual'' and ``cPurity'' function {(see Appendix~\ref{Asheep} cluster plot)}. The results showed high impurities of clusters {(Figure: \ref{sheepaxaCP})}. Two out of ten clusters were pure, both containing number ``5''. Another relatively pure cluster was composed of number ``10'' and two entries of number ``2''. The rest consisted of up to 7 different numbers. This shows that sheep counting numbers in different dialects are too different to form 10 clusters containing each number. However, considering the possibility that dialects were grouped and clustering was performed to the smaller groups, they would have reasonably pure clusters. Exploring this grouping options could be a subject for further work.
            
        \subsubsection{Linguistic and Geographical distance relationship}
        In order to investigate the correlation between linguistic and geographical distance, ``lm'' function was performed and a scatter plot was created.
        The regression line in the scatter plot suggested that the relationship existed. However, the R-squared value, extracted from the ``lm'' object, was equal to 0.131. This indicated that relationship existed, but was not significant. 
    
        One assumption made was that Cornish, Breton and Welsh dialects might have had a weakening effect on the relationship, since they had large linguistic distances compared to other dialects. However this assumption could not be validated as the correlation was less significant after eliminating them. This highlights that although these dialects had large linguistic distance scores, they also had big geographical distances that do not contradict the relationship.
        
        In addition, comparison was done between linguistic distance and \\ \(Log_{10}(\text{GeographicalDistance})\). This resulted in an even weaker relationship with R-squared being 0.097.
        
        \begin{figure}[htb]
            \centering
            \includegraphics[width=0.8\textwidth]{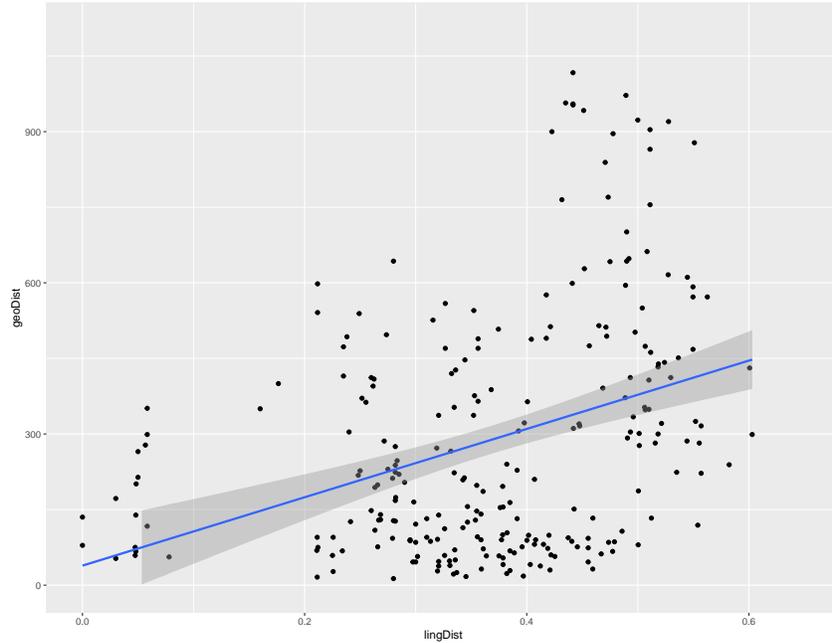}
            \caption[Linguistic and geographical distances of sheep counting systems]{Relationship between linguistic and geographical distances of sheep counting systems}
            \label{LvsG}
        \end{figure}
    
    \subsection{Colours}
    The Colours database was evaluated three different ways: getting average pairwise linguistic distance, subset pairwise linguistic distance for every word and performing all to all comparison to all groups (All languages, Indo-European, Germanic, Romance, Germanic and Romance languages). After the above mentioned data was generated, the previously defined methods were applied.
    
        \subsubsection{Mean and Standard Deviation}
\begin{figure}[htb]
            \centering
            \includegraphics[width=0.8\textwidth]{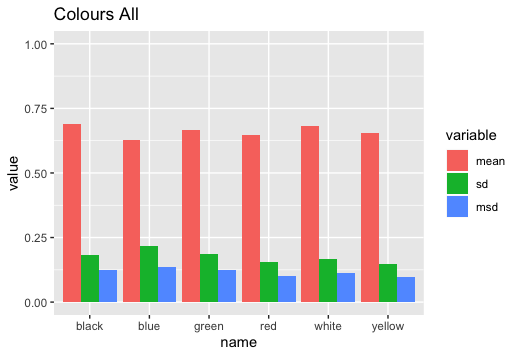}
            \caption[Statistics of ``ColoursAll'']{Mean, SD and mean*SD of every colour of all languages}
            \label{cAllmean}
        \end{figure}

        When examining the data calculated for ``ColoursAll'' none of the colours showed a clear tendency to be more preserved than others {(Figure: \ref{cAllmean})}. All colours had large distances and comparatively small standard deviation when  compared with other groups. Small standard deviation was most likely the result of most of the distances being large.
        
\begin{figure}[htb]
            \centering
            \includegraphics[width=0.8\textwidth]{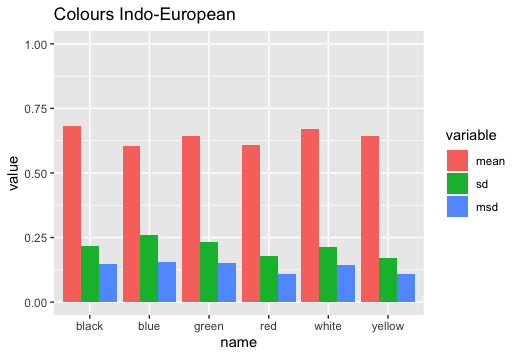}
            \caption[Statistics of Indo-European languages (colour words)]{Mean, SD and mean*SD of every colour of Indo-European languages}
            \label{cIEmean}
        \end{figure}

        Indo-European language group scores were similar to ``ColoursAll'', exhibiting slightly larger standard deviation {(Figure: \ref{cIEmean})}. Conclusion could be drawn that words for color ``Red'' are more similar in this group. The mean score of linguistic distances was 0.61, and SD was equal to 0.178, when average mean was 0.642 and SD 0.212. However, no colour stood out distinctly.
        
\begin{figure}[htb]
            \centering
            \includegraphics[width=0.8\textwidth]{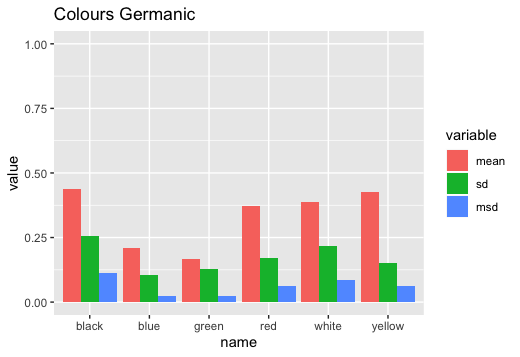}
            \caption[Statistics of Germanic languages (colour words)]{Mean, SD and mean*SD of every colour of Indo-European languages}
            \label{cGermean}
        \end{figure}

\begin{figure}[htb]
            \centering
            \includegraphics[width=0.8\textwidth]{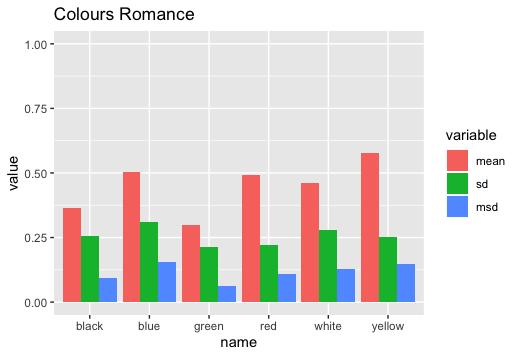}
            \caption[Statistics of Romance languages (colour words)]{Mean, SD and mean*SD of every colour of Indo-European languages}
            \label{cRommean}
        \end{figure}

        Germanic and Romance language groups revealed more significant results. Germanic languages preserved the colour ``Green'' considerably {(Figure: \ref{cGermean})}. The mean and SD was 0.168 and 0.129, when on average mean was reaching 0.333 and SD 0.171. In addition, the colour ``Blue'' had favorable scores as well - mean was 0.209 and SD was 0.106. Furthermore, Romance languages demonstrated slightly higher means and standard deviations, on average reaching 0.45 and 0.256 {(Figure: \ref{cRommean})}. Similarly to Germanic, the most preserved colour word in Romance languages was ``Green'' with a mean of 0.296 and SD of 0.214. It was followed by words for ``Black'' and then for ``Blue'', both being quite similar.
        
        \subsubsection{Density Plots}
\begin{figure}[htb]
            \centering
            \includegraphics[width=0.8\textwidth]{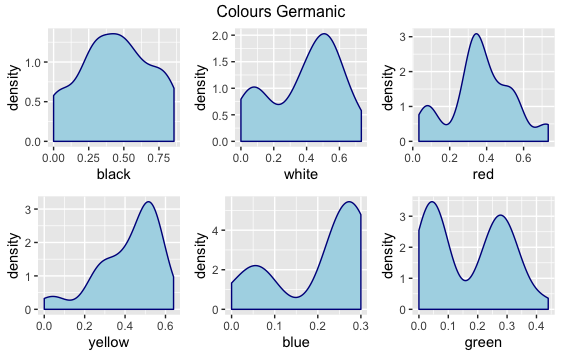}
            \caption[Density plot of Germanic languages (colour words)]{Density plots of each colour of Germanic languages}
            \label{cGerD}
        \end{figure}

\begin{figure}[htb]
            \centering
            \includegraphics[width=0.8\textwidth]{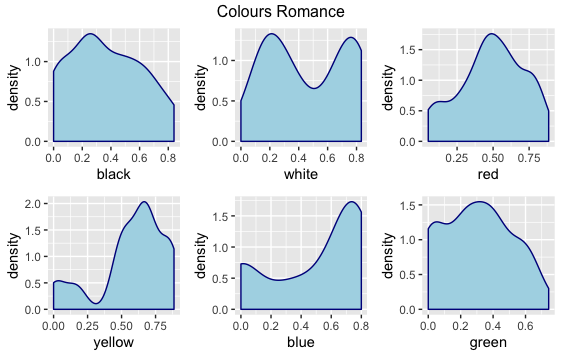}
            \caption[Density plot of Germanic languages (colour words)]{Density plots of each colour of Romance languages}
            \label{cRomD}
        \end{figure}

        Density plots of all languages and Indo-European languages were similar: both having multiple peaks with the most density around scores of 0.75 {(big linguistic distance)}. Moreover, Germanic languages density distribution consisted of two peaks for words ``White'', ``Blue'' and ``Green''  {(Figure: \ref{cGerD})}. This could possibly be the result of certain weighting in the Phonetic Substitution Table or indicate possible further grouping of languages. The color ``Black'' had more normal distribution and smoother bell shape compared to others. Furthermore, Romance languages also obtained density plots with two peaks for words ``White'', ``Yellow'', ``Blue'' {(Figure: \ref{cRomD})}. In contrast, ``Black'', ``Red'' and ``Green'' distributions were quite smooth.
        
        In order to experiment how the Phonetic Substitution Table affects the linguistic distances, ``densityP'' function was applied to the linguistic distances calculated with ``GabyTable'' substitution table. The aim was to eliminate the two peaks in the Germanic language group for word ``Green''. In Germanic languages word for green tended to begin with either ``gr'' or ``khr'' {(encoded as ``Kr'')} - both sounding similar phonetically. However, in the original substitution table, a weight for changing ``K''{(kh)} to ``g'' (and the other way around) did not exist. Consequently, a new table was implemented with this substitution. This change resulted in notably smaller linguistic distances - the mean for the word ``Green'' was 0.099. However, it did not solve the occurrence of two peaks. The density of ``Green'' again had two main peaks, but differently distributed compared to the previous case.
        
        \subsubsection{Bhattacharya Coefficients}
        Bhattacharya coefficients were calculated within each group for different pairs of colours. This helped to evaluate which colours were closer in distribution. In addition, hierarchical clustering was done with Bhattacharya coefficients {(find the dendrograms in the Appendix~\ref{Abhatt})}. However, the potential meaning behind the results was not fully examined.
        
        Another potential use of Bhattacharya coefficients is their application to the same word from a different language group. As a result, the preservation of particular words can be analysed across language groups, enabling to compare and evaluate potential reasons behind it.
        
        \subsubsection{Hierarchical Clustering}
        Hierarchical clustering with the best Silhouette value cut was performed in R for every group of formed language groups: all languages, Indo-European, Romance, Germanic, and both Germanic and Romance together. It is important to note that the results of the language group ``Romance and Germanic'' will not be discussed as it was used more for testing purposes and as expected resulted in a K=2 cut. After making the cut, one cluster consisted of Romance languages and another consisted of Germanic languages.
        
\begin{figure}[htb]
            \centering
            \includegraphics[width=0.8\textwidth]{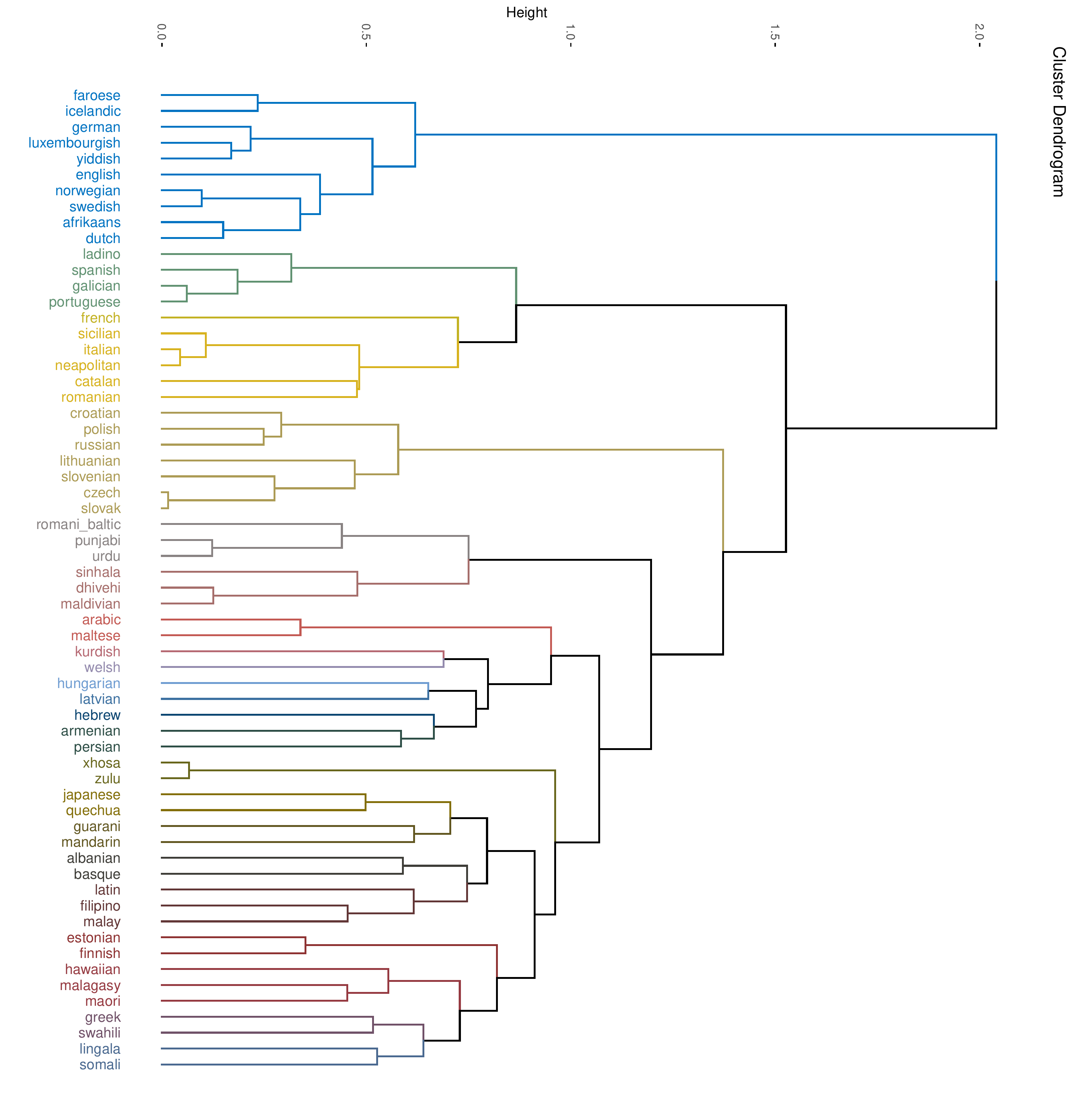}
            \caption[Dendrogram of all languages (colour words)]{Dendrogram of hierarchical clustering with Silhouette value cut for all languages (colour words)}
            \label{dendCAll}
        \end{figure}

\begin{figure}[htb]
            \centering \includegraphics[width=0.8\textwidth]{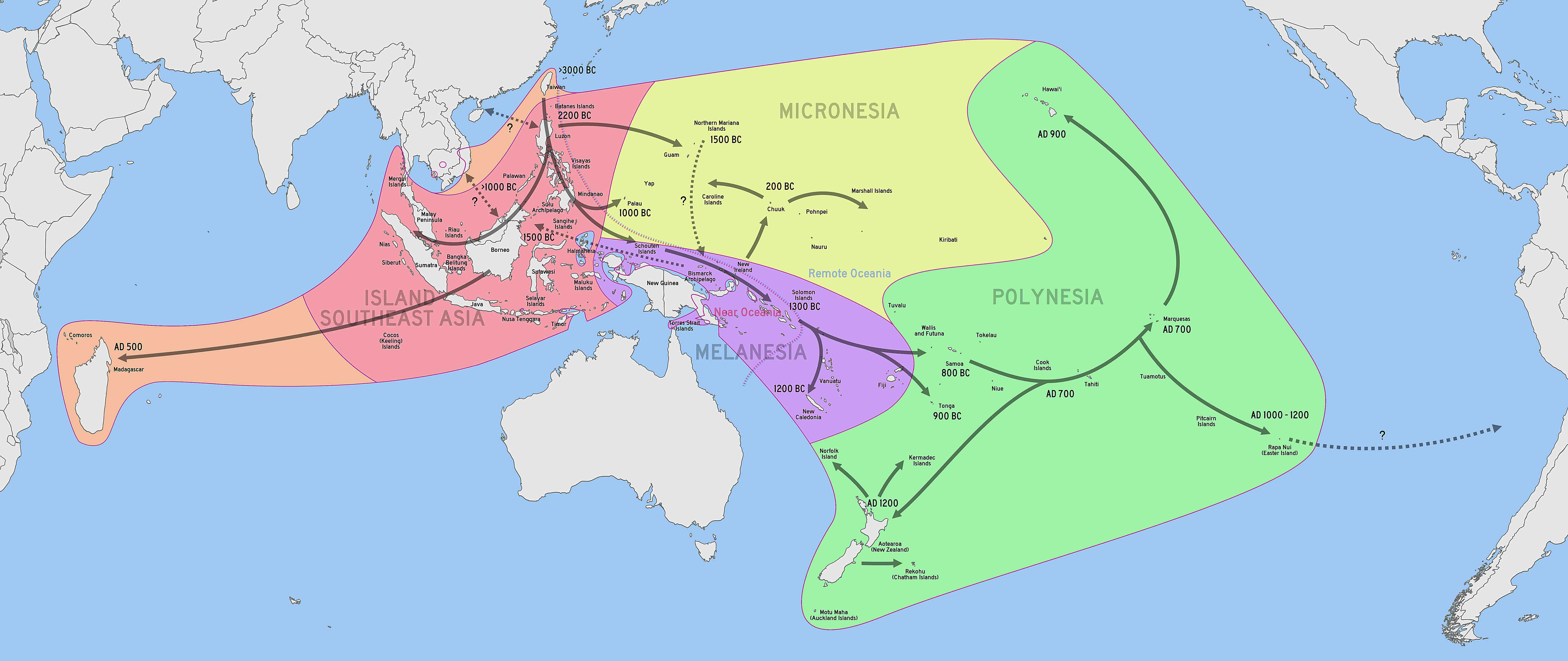}
            \caption{Chronological dispersal of the Austronesian people}
            \label{austronesian}
        \end{figure}

        To begin with, clustering of all languages showed some interesting results that complied with the grouping of the languages (find the dendrogram in Figure: \ref{dendCAll}). 
        The suggested cut by Silhouette value was 23. Some of the clusters were more a coincidence than the actual similarity of languages and did not correspond with the existing language grouping. Despite that, most of the clusters resulted in the actual language groups, or languages closely related. 
        To begin with, Baltic Romani, Punjabi' and Urdu were placed in the same cluster. Even though Baltic Romani is far away from South Asia geographically, it is believed to have originated from this area. Xhosa and Zulu formed another cluster both being the languages of the Nguni branch and spoken in South Africa. Hawaiian, Malagasy and Maori languages were grouped together and they all belong to Austronesian ethnolinguistic group \cite{austronesiaA} (see figure \ref{austronesian}).
        Sinhala (language of Sinhalese people, who make up the largest ethnic group in Sri Lanka), Dhivehi (spoken in Maldives) and Maldivian languages fell in the same group after the cut. They all are spread across islands in the Indian Ocean. Estonian and Finnish both being representatives of the Uralic language family were the same cluster. 
        
\begin{figure}[htb]
            \centering
            \includegraphics[width=0.7\textwidth]{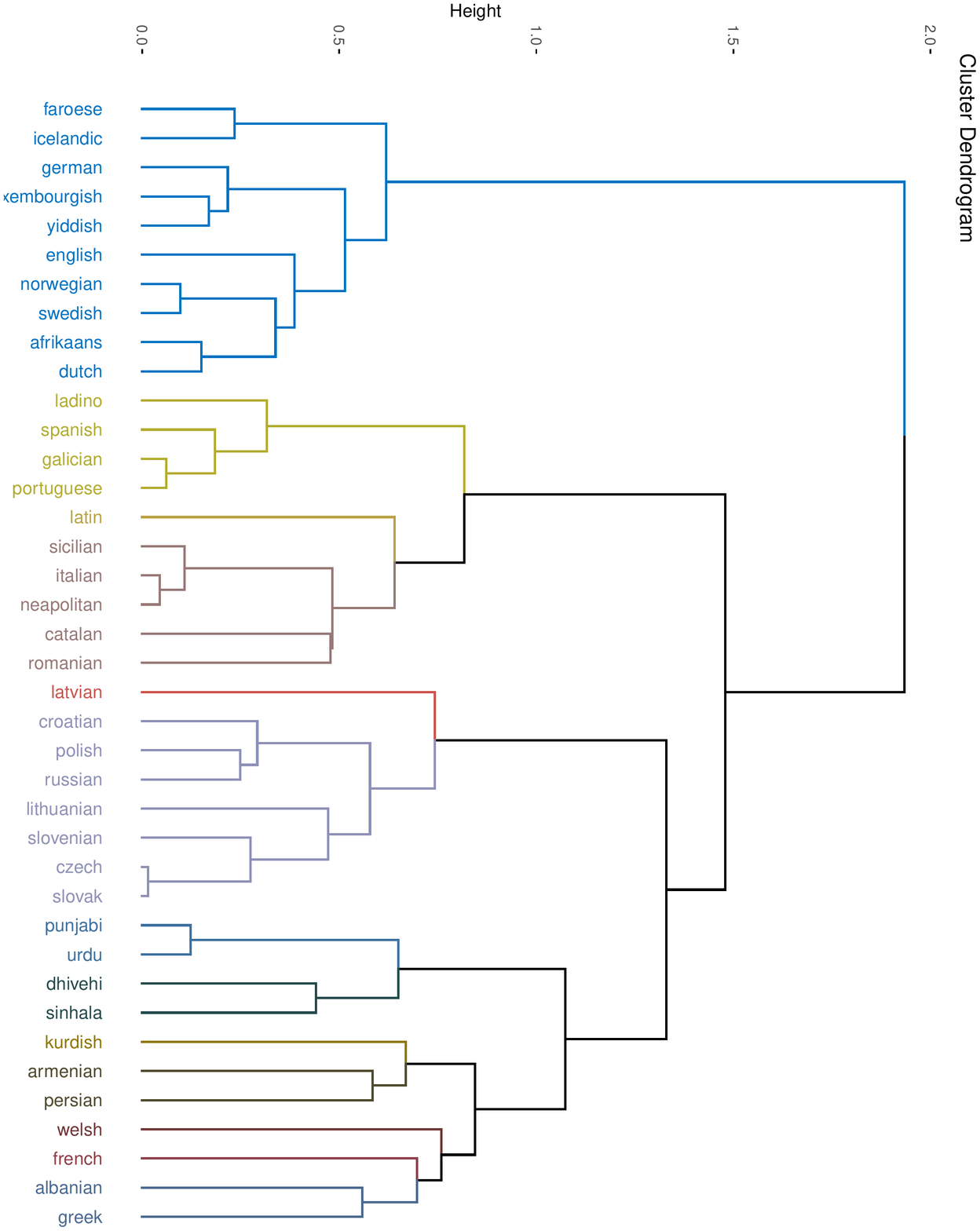}
            \caption[Dendrogram of Indo-European  languages (colour words)]{Dendrogram of hierarchical clustering with Silhouette value cut for Indo-European languages (colour words)}
            \label{dendCIE}
        \end{figure}

        Moreover, clusters of Indo-European languages were quite pure as well (groups are visible in the dendrogram of all languages, however for clarity see figure \ref{dendCIE}). There were four larger groups that stood out. First of all, the group of Germanic languages was produced accurately. It consisted of Faroese, Icelandic, German, Luxemburgish, Yiddish, English, Norwegian, Swedish, Afrikaans and Dutch. All of these languages are considered to be in the branch of Germanic languages. Another cluster was Slavic languages, which consisted of Croatian, Polish, Russian, Slovenian, Czech, Slovak and Lithuanian. Lithuanian and Latvian, according some sources, are considered to be in a separate branch, known as Baltic languages. On the other hand, in other sources they are regarded as Slavic languages. In this case, in terms of colour words Lithuanian was appointed to the Slavic languages, whereas Latvian formed a cluster on its own. In relation to Romance languages, these were divided into two groups. The first one was made of Ladino (language that derived from medieval Spanish), Spanish(Castilian), Galician and Portuguese, forming a group of the Western Romance languages. The second one consisted of Sicilian, Italian, Neapolitan, Catalan and Romanian and could be called a group of Mediterranean Romance languages. 
        
\begin{figure}[htb]
            \centering
            \includegraphics[width=0.7\textwidth]{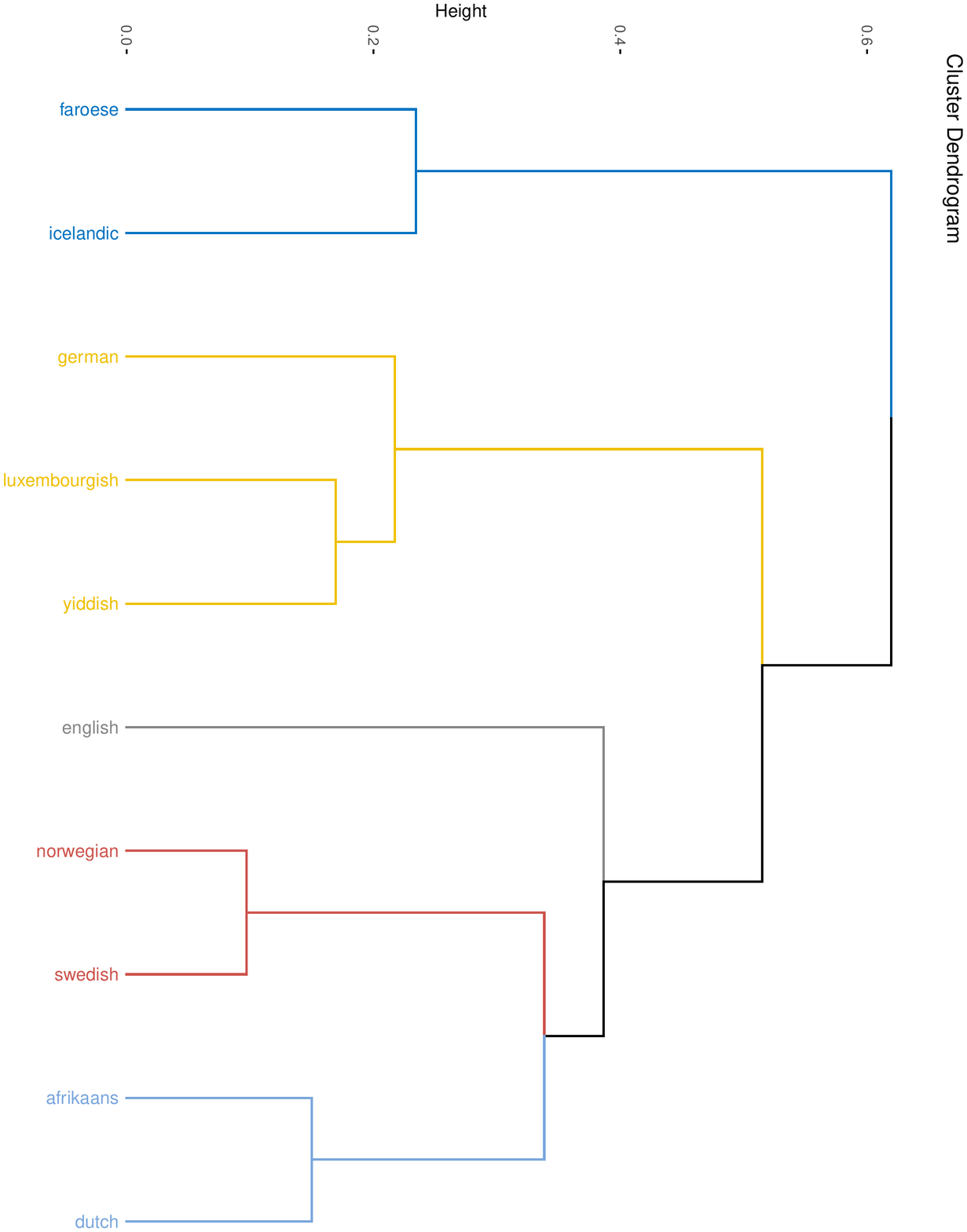}
            \caption[Dendrogram of Germanic languages (colour words) ]{Dendrogram of hierarchical clustering with Silhouette value cut for Germanic languages (colour words)}
            \label{dendG}
        \end{figure}

\begin{figure}[htb]
            \centering
            \includegraphics[width=0.7\textwidth]{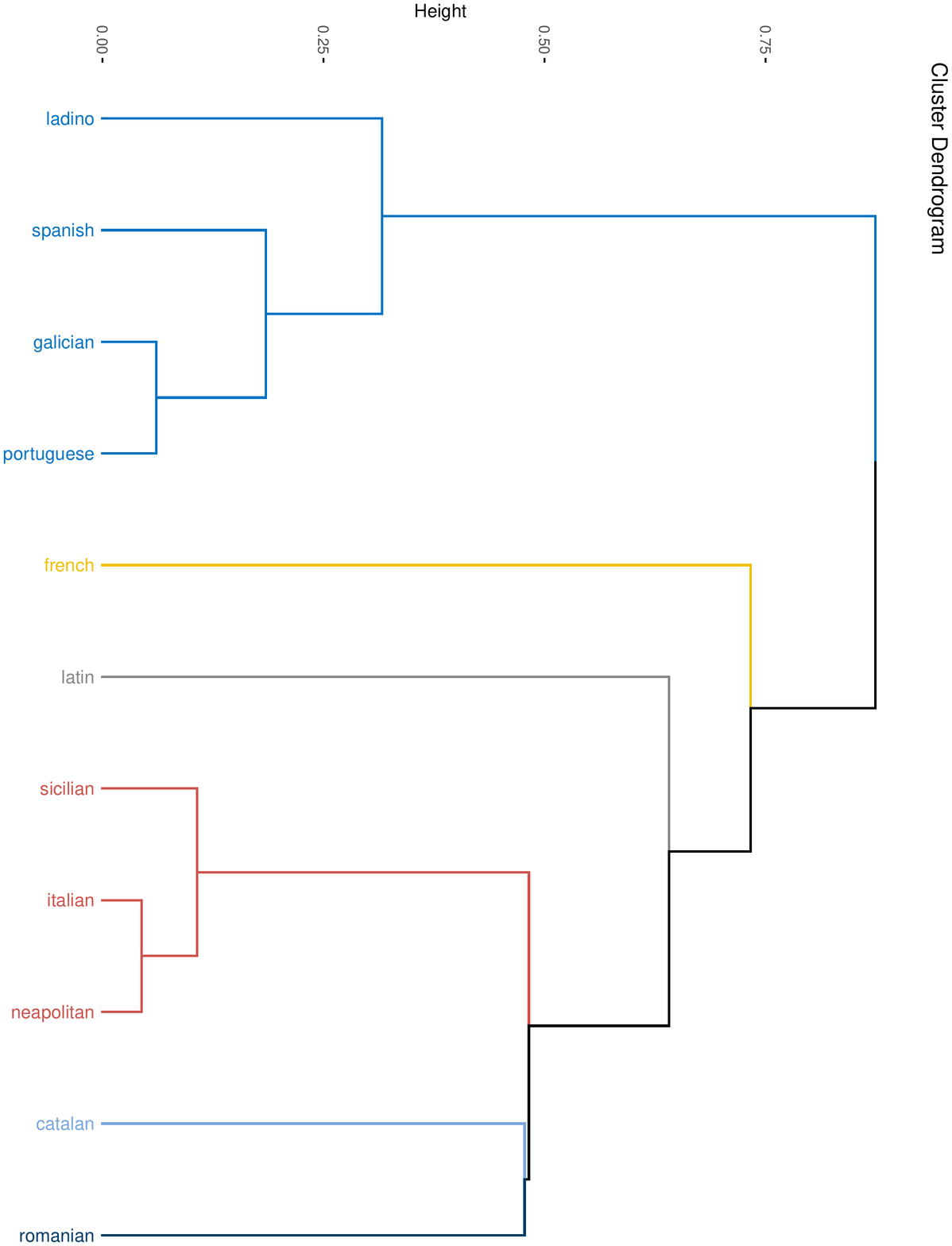}
            \caption[Dendrogram of Romance languages (colour words)]{Dendrogram of hierarchical clustering with Silhouette value cut for Romance languages (colour words)}
            \label{dendR}
        \end{figure}

        Furthermore, clustering results of the Germanic languages file (Figure: \ref{dendG}) show high relation with geographical prevalence of the languages and language development history. German, Luxembourgish (has similarities with other varieties of High German languages) and Yiddish (a High German-based language) were all in the same cluster. Also, Afrikaans and Dutch were placed in the same group, and it is known that Afrikaans derived from Dutch vernacular of South Holland in the course of 18th century. Other clusters included Faroese and Icelandic, Swedish and Norwegian, as well as English forming a cluser on its own.
        Finally, when looking at the clusters of Romance languages file (Figure \ref{dendR}) it is evident that one cluster, consisting of Ladino, Spanish, Galician and Portuguese, remained the same as in ``ColoursAll'', ``ColoursIE''. Another cluster that was formed from Romance languages in these databases was broken down into 3 clusters during separate clustering of Romance languages. Romanian and Catalan formed clusters on their own and Italian, Neopolitan and Sicilian were members of another cluster. These three languages were close geographically.
        
    \subsection{IPA}
    Hierarchical Clustering was performed to all three IPA databases and compared with the results of  hierarchical clustering of in-house phonetically encoded databases (they were created by taking subsets of German, English and Danish languages from ``Basic Words'', ``Numbers Small Collection'' and ``Colours'' databases). The first characteristic noticed was that both IPA and non-IPA databases had the same grouping of languages. This shows evidence towards substantiated phonetic encoding done in-house. Another noted tendency was that pairwise linguistic distance scores tented to be higher for IPA databases. This might be due to some graphemes  being written with a few letters in IPA databases, while phonetic encoding done in-house expressed graphemes as one symbol.
    
    Potential further work would be generating an IPA-designated Phonetic Substitution table (so far clustering has been done with ``editable'') and running the routines with the new weight table. Also, complementing the IPA databases with more languages would be an important step towards receiving more accurate results.
    
    \subsection{Small Numbers}
        \subsubsection{All to all comparison}
         Analysis was carried out in two ways.  First of all, hierarchical clustering was performed with the best silhouette value cut. For this data set best silhouette value was 0.48 and it suggested making 329 clusters. Clusters did not exhibit high purity. However, the ones that did quite clearly corresponded to unique subgroups of language families. 
         
         Another way of looking at all to all comparison data was by producing 10 clusters. The anticipated outcome was members being distinguished by numbers, forming 10 clean clusters. However, all the clusters were very impure and consisting multiple different numbers. This might be due to different languages having phonetically similar words for different words, in this case.
         
         All to all pairwise comparison could be an advantageous tool when used for language family branches or smaller, but related subsets. It could validate if languages belong to a certain group.

\section{Conclusions}
This project has aimed to develop computational methods to analyse and understand connections between human languages. 

The project included collecting words from different languages in order to form new databases, forming rules for phonetic encoding of words and adjusting phonetic substitution table. Several computational methods of calculating pairwise distance between two words were taken, including average, subset and all to all words distance calculation. It was done by incorporating edit distance and phonetic substitution table, and implementing it in SWI Prolog. This was followed by detailed analysis of distance scores, which was conducted by the specific automated routines and developed R functions. They enabled performing hierarchical clustering with a cut either according to silhouette value, or to specified K value. They provided summary of mean, standard deviation and other statistics, like Bhattacharya scores.  All these techniques delivered a thorough analysis of data and the automation of processes ensured they were used efficiently.

The resulting outcome of analysis of old sheep counting systems in different English dialects was the observation that numbers ``1'',``2'',``3'',``4'' and ``10'' were more uniform within different dialects than others, posing that they might have been the most frequently used ones. Analysis of all to all comparison did not provide pure clusters and shows that sheep counting numbers in different dialects are too different to form 10 clusters containing each number. This suggests that dialects should be grouped into subsets. Furthermore, hierarchical clustering with the best silhouette cut suggested the potential 9 groups, which consist members with the most similar counting words. Surprisingly, it was not entirely based on location. This corresponded with the difficulty of finding relationship between geographic and linguistic distance, the conducted tests showed it was insignificant.  

Analysis of colour words revealed that within Indo-European languages words for colour red were moderately more preserved. Both Germanic and Romance language groups tended to have considerably more uniform words for green and blue colours. In addition, Romance language group preserved colour black reasonably well. Analysis of linguistic distances distribution showed multiple peaks within words for various language groups, suggesting that further language grouping could be done. Furthermore, the resulting outcome of hierarchical clustering with silhouette cut was known and officially accepted language families. Most of the clusters were subgroups of existing language families. Some of them suggested different sub-grouping according to colour words (e.g. Lithuanian was appointed to Slavic languages, while Latvian formed cluster on its own).

IPA databases resulted in the same relationships between languages as non-IPA phonetically encoded databases. However, to fully explore the potential of IPA-encoded databases they ought to be expanded and a customized weights table should be created. 

In conclusion, this project resulted in creation of several felicitous computational techniques to explore many languages and their correlation all at once.

\section{Further Work}
One of the areas where further work could be performed is thorough analysis of numbers both Small and Big Collection databases, Basic words database. 

In addition, analysis routines could be enhanced by adding Bhattacharya scores, calculated in a different manner. In other words, potentially beneficial use of Bhattacharya coefficients would applying them to the same word from a different language group. As a result, the preservation of particular words could be analysed across language groups, enabling to compare and evaluate potential reasons behind it.

Moreover, regarding IPA-encoded data potential further work would be generating a customized IPA Phonetic Substitution table. Also, an important step towards receiving more accurate and interesting results would be augmenting the IPA databases with more languages.

Finally, classifying languages in language databases and automatically analysing purity of clusters would be a step forward towards fully automated and consistent process. Consequently, a list of 118 languages containing their language families and branches has been created. It could be incorporated with existing language databases.

\clearpage 
\section{Summary of contributions} 
My personal contributions during this undergraduate research assistantship include:

 \paragraph{ Data Collection.}
        \begin{itemize}
            \item Created a Sheep counting numbers database.
            \item Made geographical data database and a map of dialects.
            \item Collected colour words from 42 different languages and made a database. Made the following subsets: Indo-European, Germanic, Romance, Romance and Germanic.
            \item Created numbers, colours and basic words databases in IPA encoding.
            \item Made a list of 118 languages, their language families and branches.
        \end{itemize}
    \paragraph{ Transforming data using phonetics.} Transformed sheep counting numbers, colours (including Indo-European, Germanic, Romance, Romance and Germanic subsets) databases using a specified phonetic encoding.
    \paragraph{ Mean, SD and density analysis.} Analysed mean, SD and density of sheep numbers, colours (including all subsets). Produced tables and plots.
    \paragraph{ T-Scores and Bhattacharya calculations.} Calculated T-Scores and Bhattacharya coefficients for sheep numbers, colours (including all subsets); Made dendrograms from Bhattacharya scores.
    \paragraph{ Hierarchical clustering.}
        \begin{itemize}
            \item Performed hierarchical clustering for sheep numbers, colours (all subsets), IPA (all three). Created dendrograms.
            \item Performed hierarchical clustering with the best silhouette cut value for sheep numbers all to all, colours (all subsets), small numbers all to all. Made dendrograms, Silhouette plots, Cluster plots.
            \item Performed hierarchical clustering with k=10 cut for numbers all to all, colours (all subsets), small numbers all to all. Made dendrograms, Silhouette plots, Cluster plots.
        \end{itemize}
   \paragraph{Code development.}
        \begin{itemize}
            \item Created a package in R “CompLinguistics”, which consisted of functions: “mean\_SD”, “densityP”, “sMatrix”, “tscore”, “bhatt”, “silhouetteV”, “hcutVisual”.
            \item Produced R script that automates the processes of file reading, generating a certain format data frame, performing hierarchical clustering with the best silhouette value cut. In addition, created another R script, which performed calculations of mean, standard deviation, Bhattacharya scores and  analysis of distribution.
            \item Several shellscrips.
            \item “editableGaby” phonetic substitution table.
        \end{itemize}

\section*{Acknowledgements}
\addcontentsline{toc}{section}{Acknowledgements}

Gabija Mikulyte was supported by an undergraduate research grant from the Department of Computer Science at Brunel University London.

\newpage


\newpage
\begin{appendices}
\section{Phonetic Substitution tables}\label{table}

\subsection{Editable}\label{editable}
This table was mostly used for calculations of pairwise linguistic distances. Symbol ``\%'' indicates comments.
\begin{small}
\begin{verbatim}
%Substitution costs table
t(S1,S2,D):-
	S1=S2 -> D=0 ; 					% no cost if same character
	( t1(S1,S2,D) -> true ; ( t1(S2,S1,D) -> true ; % try S1-S2 otherwise S2-S1, in t1/3 table
		D=1)).					% else cost=1 if not in t1/3 table

% old simplistic table
%t(S1,S2,0):- S1 = S2.
%t(S1,S2,1):- S1 \== S2.

% a,b,c,d,e,f,g,h,i,j,k,l,m,n,o,p,q,r,s,t,u,v,w,x,y,z

%Grimm's law
%close consonants
t1(b,p,D):- tweight(consonant1,D). % b ->p
t1(d,t,D):- tweight(consonant1,D). % d -> t
t1(g,k,D):- tweight(consonant1,D). % g -> k
t1(p,f,D):- tweight(consonant1,D). % p -> f
t1(t,'T',D):- tweight(consonant1,D).  % t -> th
t1(k,'C',D):- tweight(consonant1,D).  % k -> ch
t1('C',h,D):- tweight(consonant1,D).  % ch -> h
t1(b,f,D):- tweight(consonant1x2,D).    % b -> p -> f
t1(d,'T',D):- tweight(consonant1x2,D).  % d -> t -> th
t1(g,'C',D):- tweight(consonant1x2,D).  % g -> k -> ch
t1(g,h,D) :- tweight(consonant1x3,D).    % g -> k -> ch -> h
t1(f,v,D):- tweight(consonant1,D).
t1(g,j,D):- tweight(consonant1,D).
t1(s,z,D):- tweight(consonant1,D).
t1(v,w,D):- tweight(consonant1,D).
t1(f,w,D):- tweight(consonant1x2,D).   % f -> v -> w
t1('F',w,D):- tweight(consonant1x2,D).   % F -> v -> w

% from numberslist10
t1(f,'F',0).  % F from ph in original
t1('S','š',0). % 'S' from sh in original
t1('C','č',0). % 'S' from sh in original
t1('T','θ',0). % 'S' from th in original

% other close consonants
t1('š',s,D):- tweight(consonant1,D).  % sh <-> s
t1('S',s,D):- tweight(consonant1,D).  % sh <-> s
t1('C','S',D):- tweight(consonant1,D). % ch <-> sh
t1('C','š',D):- tweight(consonant1,D). % ch <-> sh
t1('č','S',D):- tweight(consonant1,D). % ch <-> sh
t1('č','š',D):- tweight(consonant1,D). % ch <-> sh
t1('K',k,D):- tweight(consonant1,D).  % kh <-> k
t1('G',k,D):- tweight(consonant1,D).  % gh <-> k
t1('G',g,D):- tweight(consonant1,D).  % gh <-> g
t1('K','G',D):- tweight(consonant1,D).  % kh <->gh
t1('Z',z,D):- tweight(consonant1,D).  % zh <-> z
t1(c,s,D):- tweight(consonant1,D).    % ts <-> s
t1(x,k,D):- tweight(consonant1,D).    % ks <-> k
t1('D',d,D):-tweight(consonant1,D).    % dh <-> d
% vowels
%t1(S1,S2,0.2):- Vowels=[a,e,i,o,u,y], member(S1,Vowels), member(S2,Vowels).
t1(a,Y,V):- (Y=e;Y='E';Y=i;Y='I';Y=o;Y='O';Y=u;Y='U';Y=y;Y='Y'), tweight(vowel,V).
t1(e,Y,V):- (Y=a;Y='A';Y=i;Y='I';Y=o;Y='O';Y=u;Y='U';Y=y;Y='Y'), tweight(vowel,V).
t1(i,Y,V):- (Y=a;Y='A';Y=e;Y='E';Y=o;Y='O';Y=u;Y='U';Y=y;Y='Y'), tweight(vowel,V).
t1(o,Y,V):- (Y=a;Y='A';Y=e;Y='E';Y=i;Y='I';Y=u;Y='U';Y=y;Y='Y'), tweight(vowel,V).
t1(u,Y,V):- (Y=a;Y='A';Y=e;Y='E';Y=i;Y='I';Y=o;Y='O';Y=y;Y='Y'), tweight(vowel,V).
t1(y,Y,V):- (Y=a;Y='A';Y=e;Y='E';Y=i;Y='I';Y=o;Y='O';Y=u;Y='U'), tweight(vowel,V).

% same vowel
t1(A1,A2,0):- t_a(A1), t_a(A2).
t1(E1,E2,0):- t_e(E1), t_e(E2).
t1(I1,I2,0):- t_i(I1), t_i(I2).
t1(O1,O2,0):- t_o(O1), t_o(O2).
t1(U1,U2,0):- t_u(U1), t_u(U2).
t1(Y1,Y2,0):- t_y(Y1), t_y(Y2).

% close vowels
t1(X,Y,V):- tvowel(X), tvowel(Y), tweight(vowel,V).

% long vowels
t1('A',Y,V):- (Y='E';Y=e;Y='I';Y=i;Y='O';Y=o;Y='U';Y=u;Y='Y';Y=y), tweight(vowel,V).
t1('E',Y,V):- (Y='A';Y=a;Y='I';Y=i;Y='O';Y=o;Y='U';Y=u;Y='Y';Y=y), tweight(vowel,V).
t1('I',Y,V):- (Y='A';Y=a;Y='E';Y=e;Y='O';Y=o;Y='U';Y=u;Y='Y';Y=y), tweight(vowel,V).
t1('O',Y,V):- (Y='A';Y=a;Y='E';Y=e;Y='I';Y=i;Y='U';Y=u;Y='Y';Y=y), tweight(vowel,V).
t1('U',Y,V):- (Y='A';Y=a;Y='E';Y=e;Y='I';Y=i;Y='O';Y=o;Y='Y';Y=y), tweight(vowel,V).
t1('Y',Y,V):- (Y='A';Y=a;Y='E';Y=e;Y='I';Y=i;Y='O';Y=o;Y='U';Y=u), tweight(vowel,V).


%long-short vowels
t1('A',a,Z):- tweight(longvowel,Z).
t1('E',e,Z):- tweight(longvowel,Z).
t1('I',i,Z):- tweight(longvowel,Z).
t1('O',o,Z):- tweight(longvowel,Z).
t1('U',u,Z):- tweight(longvowel,Z).
t1('Y',y,Z):- tweight(longvowel,Z).

%long consonants
t1('M',m,Z):- tweight(longconsonant,Z).
t1('N',n,Z):- tweight(longconsonant,Z).

% weight table
tweight(vowel,0.2).
tweight(longvowel,0.1).
tweight(consonant1,0.2).
tweight(consonant1x2,0.4).
tweight(consonant1x3,0.8).
tweight(longconsonant,0.05).

tvowel(V):- t_a(V); t_e(V); t_i(V); t_o(V); t_u(V); t_y(V).
\end{verbatim}
\end{small}

\subsection{editableGaby}\label{gaby}
This table was created based on Editable illustrated before. Comments and ``!!'' symbol indicates where changes were made.
\begin{small}
\begin{verbatim}
% Substitution costs table
t(S1,S2,D):-
	S1=S2 -> D=0 ; 					% no cost if same character
	( t1(S1,S2,D) -> true ; ( t1(S2,S1,D) -> true ; % try S1-S2 otherwise S2-S1, in t1/3 table
		D=1)).					% else cost=1 if not in t1/3 table

% old simplistic table
%t(S1,S2,0):- S1 = S2.
%t(S1,S2,1):- S1 \== S2.

% a,b,c,d,e,f,g,h,i,j,k,l,m,n,o,p,q,r,s,t,u,v,w,x,y,z

/*
Phonetic encoding
c - ts
x - ks
C - ch as in charity
k - as in cat
T - th
S - sh
G - dzh                              %!!
K - kh
Z - zh
D - dz
H - spanish/portuguese sound  of 'j' %!! 
F - ph
A,I,O,U,Y - long vowels
*/

% Grimm's law
%close consonants
t1(b,p,D):- tweight(consonant1,D). % b ->p
t1(d,t,D):- tweight(consonant1,D). % d -> t
t1(g,k,D):- tweight(consonant1,D). % g -> k
t1(p,f,D):- tweight(consonant1,D). % p -> f
t1(t,'T',D):- tweight(consonant1,D).  % t -> th
t1(k,'C',D):- tweight(consonant1x2,D).  % k -> ch  !!    
t1('C',h,D):- tweight(consonant1x2,D).  % ch -> h  !!    
t1(b,f,D):- tweight(consonant1x2,D).    % b -> p -> f
t1(d,'T',D):- tweight(consonant1x2,D).  % d -> t -> th
t1(g,'C',D):- tweight(consonant1x2,D).  % g -> ch        
t1(g,h,D) :- tweight(consonant1x1,D).   %g->k & g->h & k->h same, ch further  !!
t1(f,v,D):- tweight(consonant1,D).
t1(g,j,D):- tweight(consonant1,D).      %!!
t1(s,z,D):- tweight(consonant1,D).
t1(v,w,D):- tweight(consonant1,D).
t1(f,w,D):- tweight(consonant1x2,D).     % f -> v -> w
t1('F',w,D):- tweight(consonant1x2,D).   % F -> v -> w

% from numberslist10
t1(f,'F',0).  % F from ph in original
t1('S','š',0). % 'S' from sh in original
t1('C','č',0). % 'S' from sh in original
t1('T','θ',0). % 'S' from th in original

% other close consonents
t1('š',s,D):- tweight(consonant1,D).  % sh <-> s
t1('S',s,D):- tweight(consonant1,D).  % sh <-> s
t1('C','S',D):- tweight(consonant1,D). % ch <-> sh
t1('C','š',D):- tweight(consonant1,D). % ch <-> sh
t1('č','S',D):- tweight(consonant1,D). % ch <-> sh
t1('č','š',D):- tweight(consonant1,D). % ch <-> sh
t1('K',k,D):- tweight(consonant1,D).  % kh <-> k
t1('K',g,D):- tweight(consonant1,D).  % kh <-> g
t1('G','Z',D):- tweight(consonant1,D).  % dzh <-> zh !!
t1('G','C',D):- tweight(consonant1,D).  % dzh <-> ch !!
t1('K','G',D):- tweight(consonant1,D).  % kh <->gh
t1('Z',z,D):- tweight(consonant1,D).  % zh <-> z
t1('Z',s,D):- tweight(consonant1x2,D).  % zh -> z -> s !!
t1(c,s,D):- tweight(consonant1,D).    % ts <-> s
t1(x,k,D):- tweight(consonant1,D).    % ks <-> k
t1('D',d,D):-tweight(consonant1,D).    % dh <-> d
t1('K',g,D):-tweight(consonant1,D).  % kh -> g		%!!
t1('H','K',D):-tweight(consonant1,D).			%!!
t1('H',g,D):-tweight(consonant1,D).			%!!
t1('H',k,D):-tweight(consonant1,D).			%!!
t1('H',h,D):-tweight(consonant1,D).			%!!

% vowels
%t1(S1,S2,0.2):- Vowels=[a,e,i,o,u,y], member(S1,Vowels), member(S2,Vowels).
t1(a,Y,V):- (Y=e;Y='E';Y=i;Y='I';Y=o;Y='O';Y=u;Y='U';Y=y;Y='Y'), tweight(vowel,V).
t1(e,Y,V):- (Y=a;Y='A';Y=i;Y='I';Y=o;Y='O';Y=u;Y='U';Y=y;Y='Y'), tweight(vowel,V).
t1(i,Y,V):- (Y=a;Y='A';Y=e;Y='E';Y=o;Y='O';Y=u;Y='U';Y=y;Y='Y'), tweight(vowel,V).
t1(o,Y,V):- (Y=a;Y='A';Y=e;Y='E';Y=i;Y='I';Y=u;Y='U';Y=y;Y='Y'), tweight(vowel,V).
t1(u,Y,V):- (Y=a;Y='A';Y=e;Y='E';Y=i;Y='I';Y=o;Y='O';Y=y;Y='Y'), tweight(vowel,V).
t1(y,Y,V):- (Y=a;Y='A';Y=e;Y='E';Y=i;Y='I';Y=o;Y='O';Y=u;Y='U'), tweight(vowel,V).

% same vowel
t1(A1,A2,0):- t_a(A1), t_a(A2).
t1(E1,E2,0):- t_e(E1), t_e(E2).
t1(I1,I2,0):- t_i(I1), t_i(I2).
t1(O1,O2,0):- t_o(O1), t_o(O2).
t1(U1,U2,0):- t_u(U1), t_u(U2).
t1(Y1,Y2,0):- t_y(Y1), t_y(Y2).

% close vowels
t1(X,Y,V):- tvowel(X), tvowel(Y), tweight(vowel,V).

% long vowels
t1('A',Y,V):- (Y='E';Y=e;Y='I';Y=i;Y='O';Y=o;Y='U';Y=u;Y='Y';Y=y), tweight(vowel,V).
t1('E',Y,V):- (Y='A';Y=a;Y='I';Y=i;Y='O';Y=o;Y='U';Y=u;Y='Y';Y=y), tweight(vowel,V).
t1('I',Y,V):- (Y='A';Y=a;Y='E';Y=e;Y='O';Y=o;Y='U';Y=u;Y='Y';Y=y), tweight(vowel,V).
t1('O',Y,V):- (Y='A';Y=a;Y='E';Y=e;Y='I';Y=i;Y='U';Y=u;Y='Y';Y=y), tweight(vowel,V).
t1('U',Y,V):- (Y='A';Y=a;Y='E';Y=e;Y='I';Y=i;Y='O';Y=o;Y='Y';Y=y), tweight(vowel,V).
t1('Y',Y,V):- (Y='A';Y=a;Y='E';Y=e;Y='I';Y=i;Y='O';Y=o;Y='U';Y=u), tweight(vowel,V).


%long-short vowels
t1('A',a,Z):- tweight(longvowel,Z).
t1('E',e,Z):- tweight(longvowel,Z).
t1('I',i,Z):- tweight(longvowel,Z).
t1('O',o,Z):- tweight(longvowel,Z).
t1('U',u,Z):- tweight(longvowel,Z).
t1('Y',y,Z):- tweight(longvowel,Z).

%long consonants
t1('M',m,Z):- tweight(longconsonant,Z).
t1('N',n,Z):- tweight(longconsonant,Z).

% weight table
tweight(vowel,0.2).
tweight(longvowel,0.1).
tweight(consonant1,0.2).
tweight(consonant1x2,0.4).
tweight(consonant1x3,0.8).
tweight(longconsonant,0.05).

tvowel(V):- t_a(V); t_e(V); t_i(V); t_o(V); t_u(V); t_y(V).
\end{verbatim}
\end{small}

\section{Dendrograms and Cluster plots}
\subsection{Sheep counting systems}\label{Asheep}
\begin{figure}
    \centering
    \includegraphics[width=0.7\textwidth]{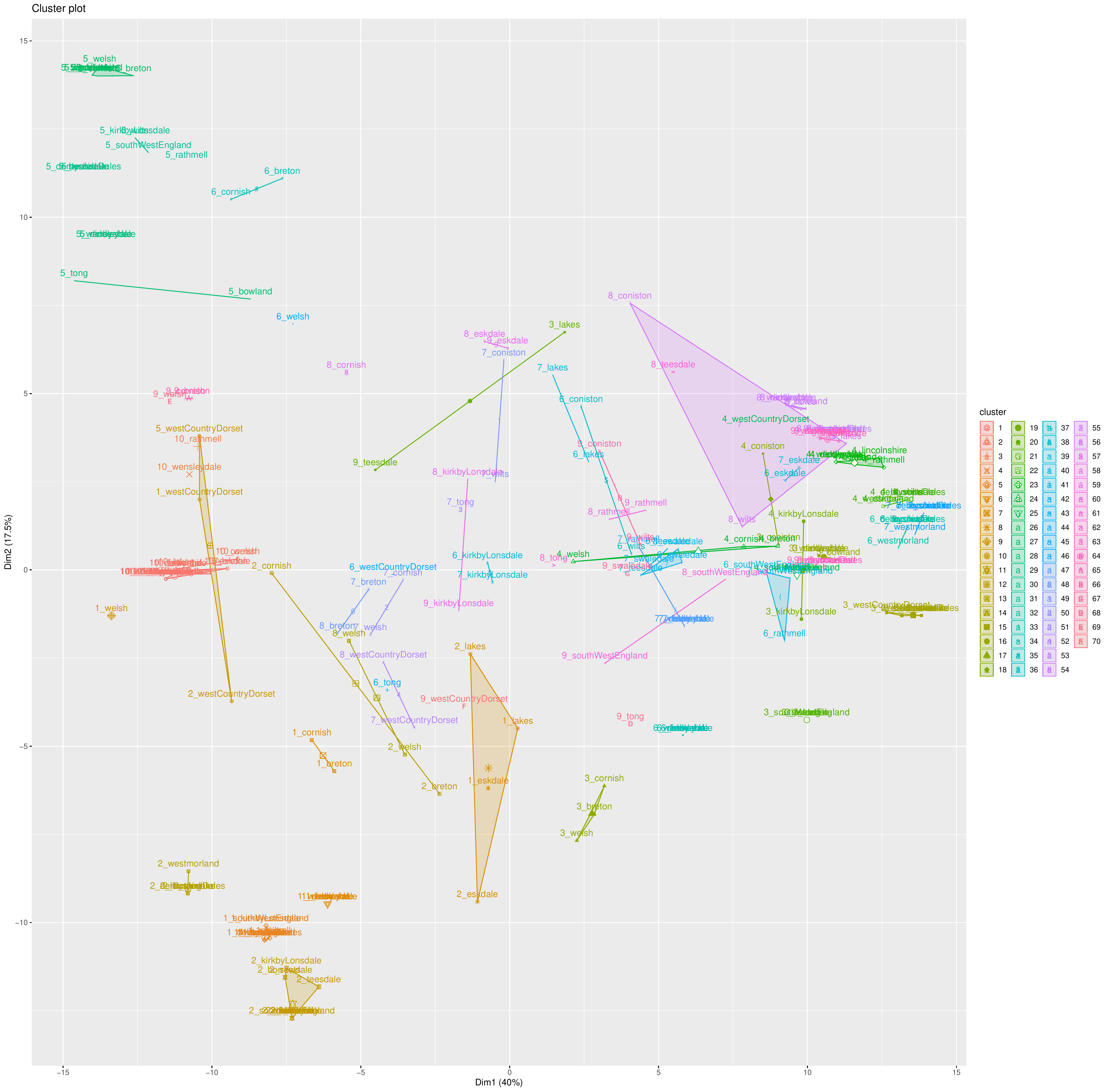}
    \caption[Cluster plot of sheep counting all to all comparison (k=10)]{Cluster plot of all to all sheep counting systems comparison with best silhouette value cut}
    \label{sheepCP}
\end{figure}

\begin{figure}
    \centering
    \includegraphics[width=0.7\textwidth]{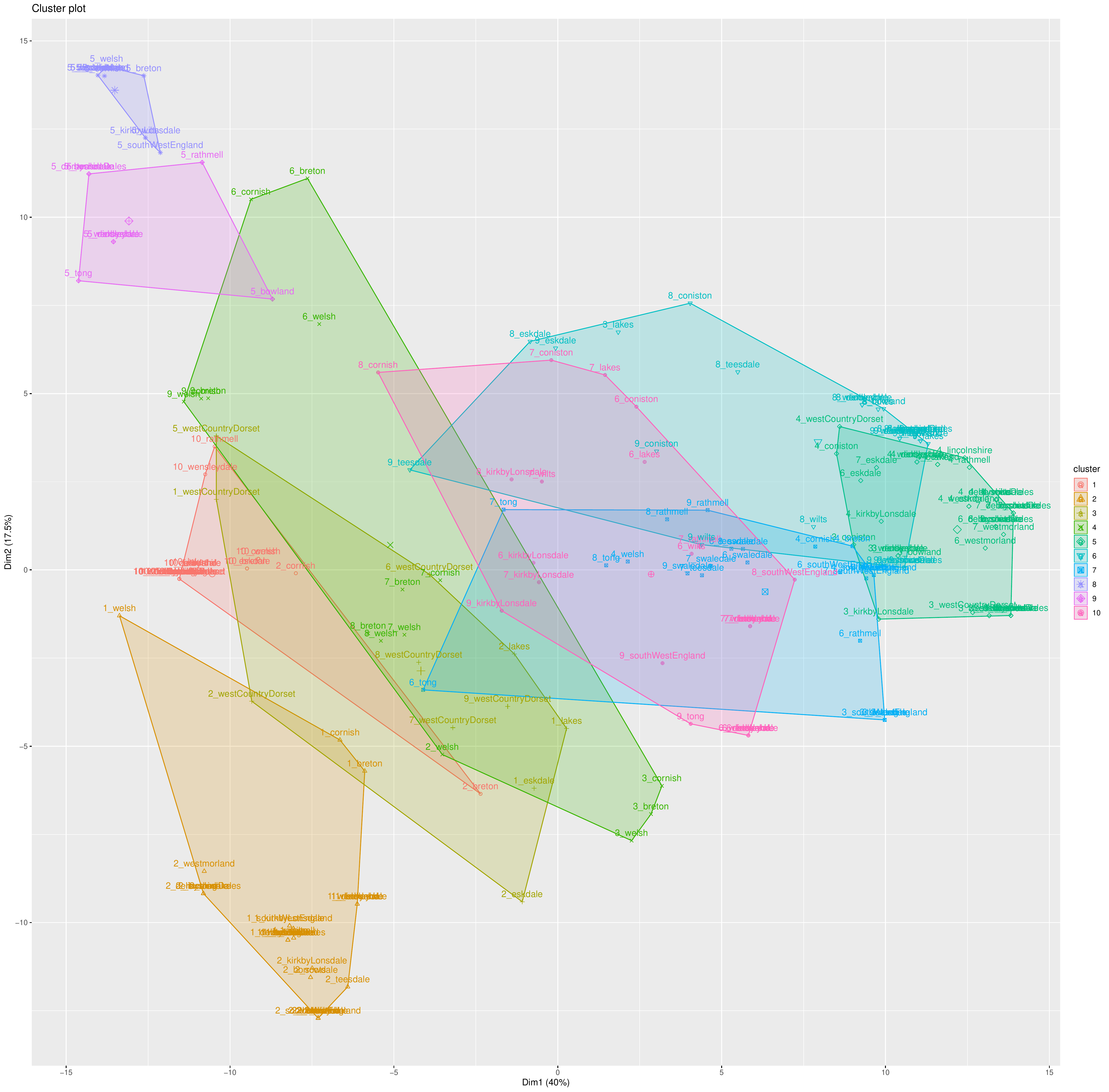}
    \caption[Cluster plot of sheep counting all to all comparison (k=10)]{Cluster plot of all to all sheep counting systems comparison with K=10 cut}
    \label{sheepCPk10}
\end{figure}

Figures \ref{sheepCP} and \ref{sheepCPk10}.

\subsection{Dendrograms of Bhattacharya scores of colour words}\label{Abhatt}
\begin{figure}
    \centering
    \includegraphics[width=0.7\textwidth]{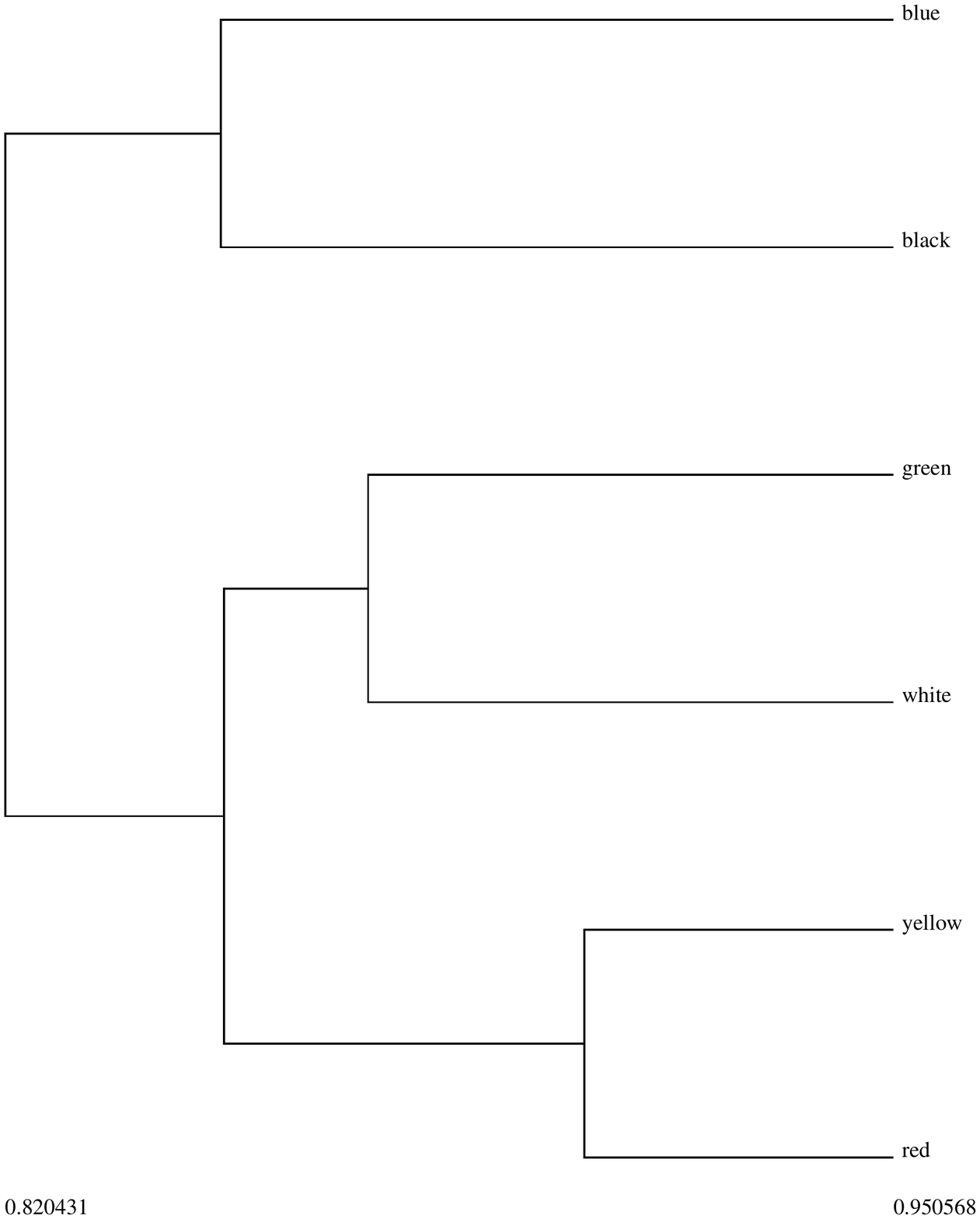}
    \caption{Dendrogram of Bhattacharya scores of ``ColoursAll''}
    \label{bhattCall}
\end{figure}

\begin{figure}
    \centering
    \includegraphics[width=0.7\textwidth]{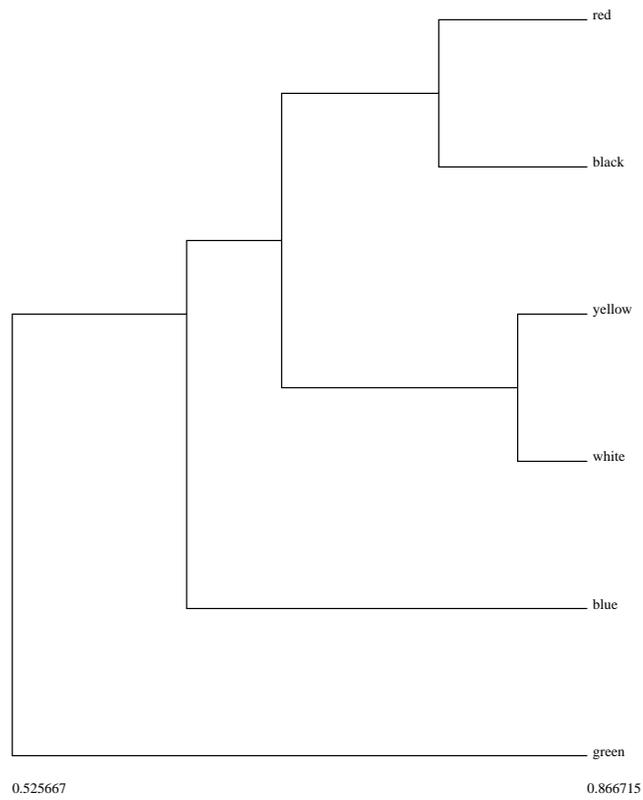}
    \caption{Dendrogram of Bhattacharya scores of Indo-European languages (colours)}
    \label{bhattCIE}
\end{figure}

\begin{figure}
    \centering
    \includegraphics[width=0.7\textwidth]{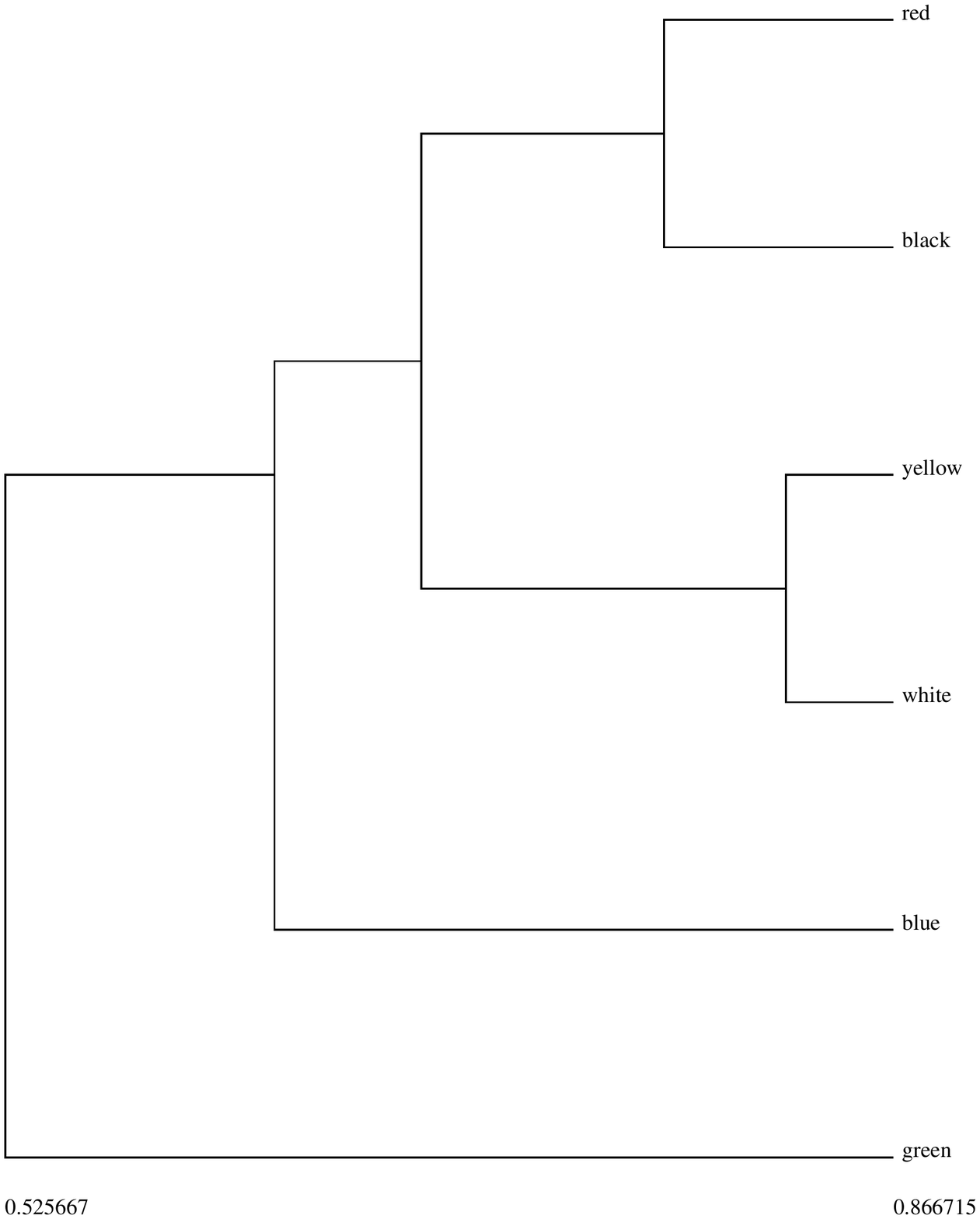}
    \caption{Dendrogram of Bhattacharya scores of Germanic languages (colours)}
    \label{bhattCG}
\end{figure}

\begin{figure}
    \centering
    \includegraphics[width=0.7\textwidth]{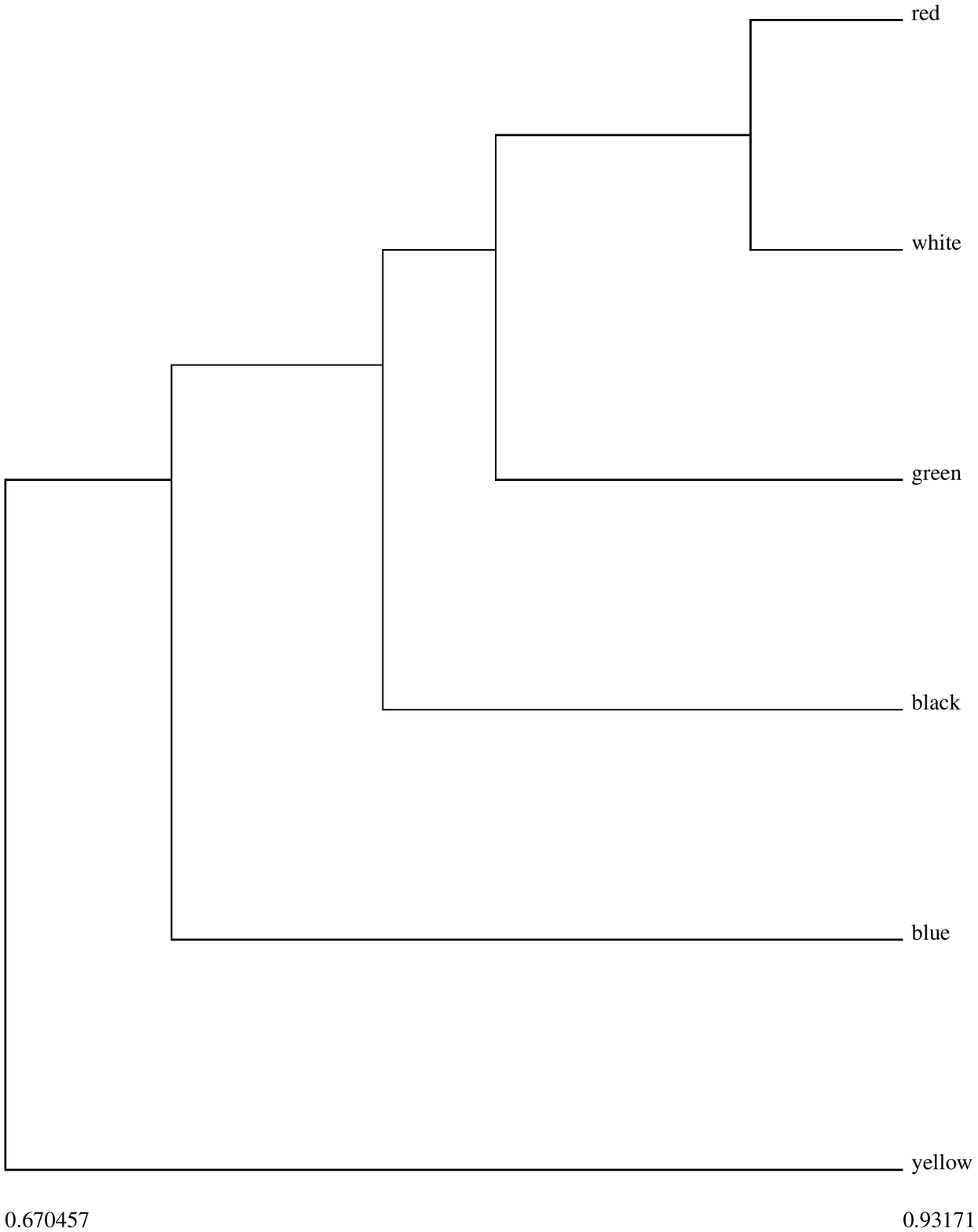}
    \caption{Dendrogram of Bhattacharya scores of Romance languages (colours)}
    \label{bhattCR}
\end{figure}

\begin{figure}
    \centering
    \includegraphics[width=0.7\textwidth]{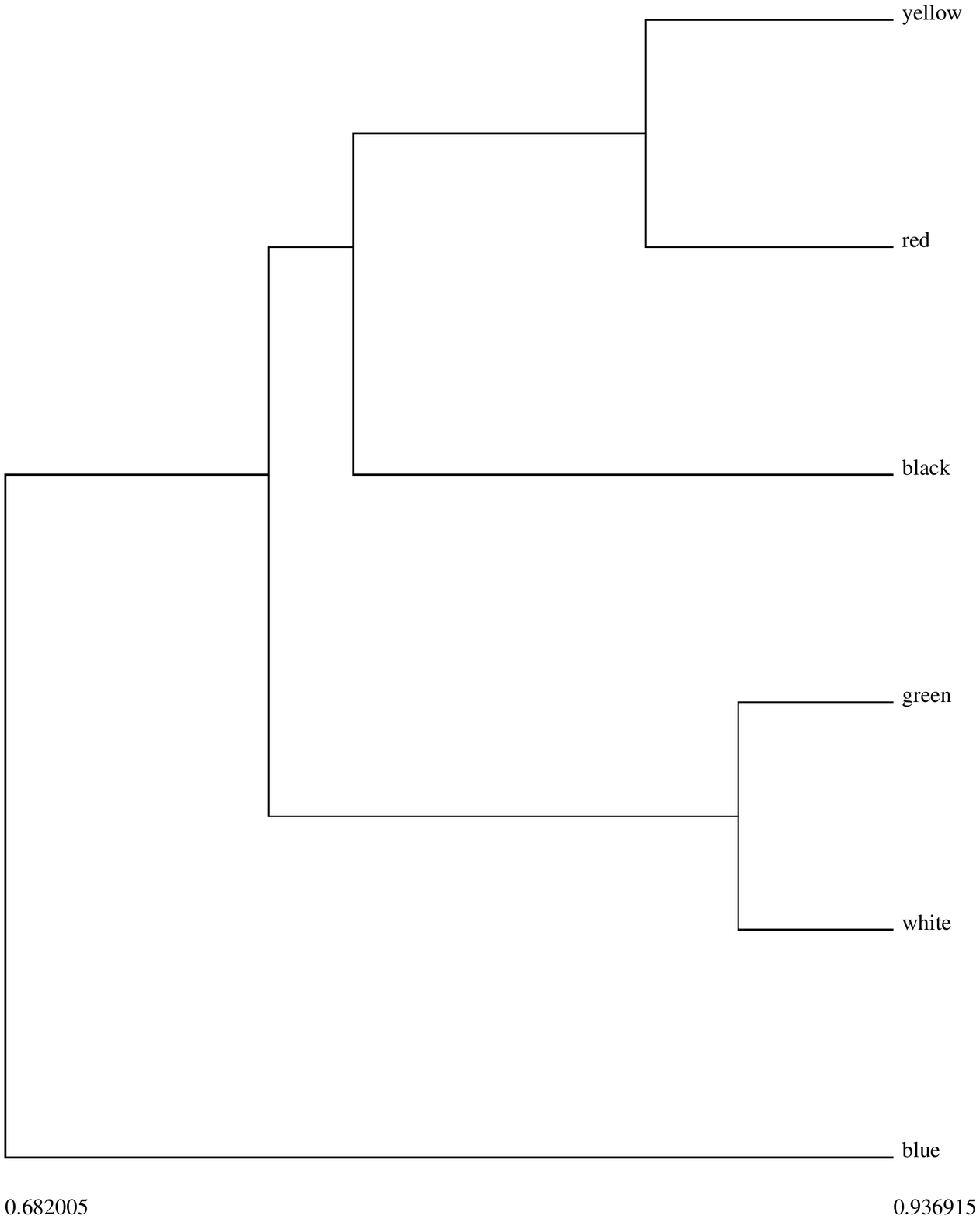}
    \caption{Dendrogram of Bhattacharya scores of Germanic and Romance languages (colours)}
    \label{bhattCGR}
\end{figure}

Figures \ref{bhattCall}, \ref{bhattCIE}, \ref{bhattCG}, \ref{bhattCR} and \ref{bhattCGR}.

\end{appendices}

\end{document}